\documentclass{article}

\usepackage[nonatbib,final]{neurips_2020}

\usepackage[utf8]{inputenc} % allow utf-8 input
\usepackage[T1]{fontenc}    % use 8-bit T1 fonts
\usepackage{hyperref}       % hyperlinks
\usepackage{url}            % simple URL typesetting
\usepackage{booktabs}       % professional-quality tables
\usepackage{amsfonts}       % blackboard math symbols
\usepackage{nicefrac}       % compact symbols for 1/2, etc.
\usepackage{microtype}
\usepackage{graphicx}
\usepackage{subfigure}
\usepackage[svgnames]{xcolor}
\usepackage{algorithm2e}
\usepackage{algorithm}
\usepackage[noend]{algorithmic}

\usepackage{hyperref}

\usepackage{color}
\usepackage{varwidth}
\usepackage{xspace}

\usepackage{amsmath,amsfonts,bm}

\def\eqref#1{equation~\ref{#1}}

\def\ceil#1{\lceil #1 \rceil}

\def\1{\bm{1}}

\DeclareMathAlphabet{\mathsfit}{\encodingdefault}{\sfdefault}{m}{sl}
\SetMathAlphabet{\mathsfit}{bold}{\encodingdefault}{\sfdefault}{bx}{n}

\def\gO{{\mathcal{O}}}

\newcommand{\todo}[1]{ \textcolor{red}{\bf #1}}

\definecolor{clr}{cmyk}{0, 0.7808, 0.4429, 0.1412}

\renewcommand{\todo}[1]{}

\usepackage{mathtools}
\usepackage{amsmath,amssymb}
\usepackage{bm}
\usepackage{siunitx}
\usepackage{wrapfig}
\newcommand{\E}{{\mathbb{E}}}
\newcommand{\pp}{pick\&place\xspace}

\newtheorem{prop}{Proposition}

\makeatletter
\def\blfootnote{\xdef\@thefnmark{}\@footnotetext}
\makeatother

\usepackage[numbers]{natbib}
 \author{
	Karl Pertsch$^{*,1}$ \hspace*{16pt} Oleh Rybkin$^{*, 2}$ \hspace*{16pt} Frederik Ebert$^3$ \\[0.1cm]
	\textbf{ Chelsea Finn$^4$ \hspace*{16pt} Dinesh Jayaraman$^2$ \hspace*{16pt}  Sergey Levine$^3$ }\\[0.2cm]
	$^1$ USC \hspace*{20pt}  $^2$ UPenn \hspace*{20pt} $^3$ UC Berkeley \hspace*{20pt}  $^4$ Stanford University
}

\title{Long-Horizon Visual Planning \\ with Goal-Conditioned Hierarchical Predictors}
\begin{document}
\maketitle
\blfootnote{\footnotesize{$^*$ Equal contribution. Ordering determined by a coin flip. Project page: \url{orybkin.github.io/video-gcp}}}




\begin{abstract}

The ability to predict and plan into the future is fundamental for agents acting in the world. To reach a faraway goal, we predict trajectories at multiple timescales, first devising a coarse plan towards the goal and then gradually filling in details. In contrast, current learning approaches for visual prediction and planning fail on long-horizon tasks as they generate predictions (1)~without considering goal information, and (2)~at the finest temporal resolution, one step at a time. In this work we propose a framework for visual prediction and planning that is able to overcome both of these limitations. First, we formulate the problem of predicting \emph{towards a goal} and propose the corresponding class of latent space goal-conditioned predictors (GCPs). GCPs significantly improve planning efficiency by constraining the search space to only those trajectories that reach the goal. Further, we show how GCPs can be naturally formulated as hierarchical models that, given two observations, predict an observation between them, and by recursively subdividing each part of the trajectory generate complete sequences. This divide-and-conquer strategy is effective at long-term prediction, and enables us to design an effective hierarchical planning algorithm that optimizes trajectories in a coarse-to-fine manner. We show that by using both goal-conditioning and hierarchical prediction, GCPs enable us to solve visual planning tasks with much longer horizon than previously possible.%\footnote{See prediction and planning videos on the supplementary website: }

%The ability to predict and plan into the future is fundamental for agents acting in the world. Standard approaches for doing so entail generating predictions at a fixed frequency, for each discrete time step in turn. Instead, we propose to perform these operations starting from a higher level of abstraction, generating events hierarchically between a given start and end observation. To this end, we propose Hierarchical event-driven generation (HEDGE). HEDGE models long state sequences, such as videos, by hierarchically decomposing them in time and recursively filling in finer and finer details at each level of the hierarchy. Its fundamental operator seeks to answer: given two states, what is one event that occurs between them? This operator, applied recursively, can generate complete sequences. We formulate a graphical model to represent recursion trees corresponding to this generative process, instantiate it with deep neural networks, and apply amortized variational inference to infer the latent tree structure from data. At test-time, our model can fill in predictions between a given start and goal frame recursively and in a time-agnostic manner. In one simulated and one real video dataset, we find that HEDGE maintains consistent predictions over long horizons, outperforming standard forward prediction methods that predict left-to-right.

\end{abstract}

\section{Introduction}

\begin{wrapfigure}{R}{0.6\textwidth}
    \centering
    \vspace{-0.16in}
    \includegraphics[width=1\linewidth]{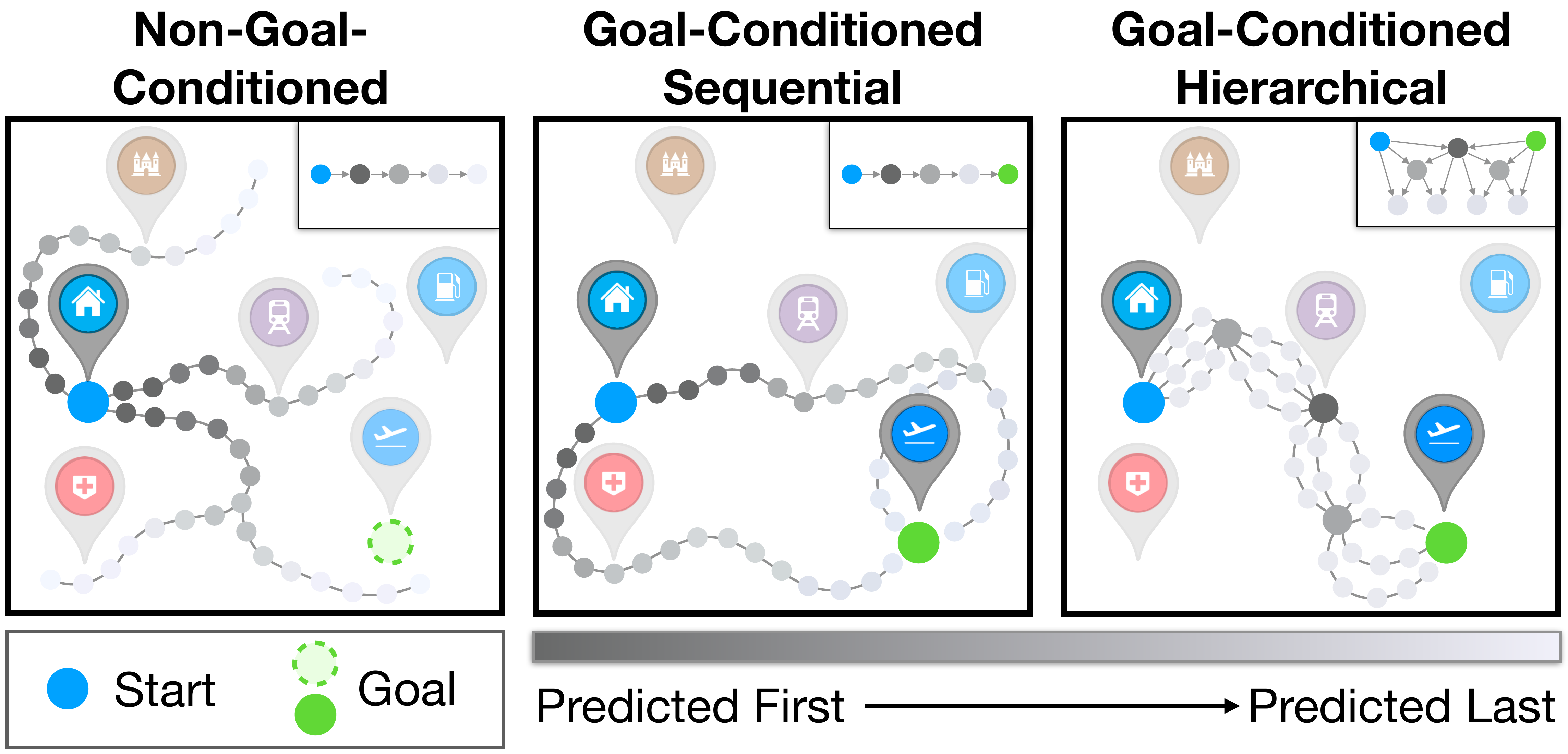}
    \vskip -2mm
    \caption{%
    When planning towards faraway goals, we propose to condition the prediction of candidate trajectories on the goal, which significantly reduces the search space of possible trajectories (\textbf{left} vs. \textbf{middle}) and enables hierarchical planning approaches that break a long-horizon task into a series of short-horizon tasks by placing subgoals (\textbf{right}).
    }
    \label{fig:teaser}
    \vspace{-0.29in}
\end{wrapfigure}

Intelligent agents aiming to solve long-horizon tasks reason about the future, make predictions, and plan accordingly. 
%Equipping artificial agents with expressive predictive models holds the promise of enabling them to learn such long-horizon problem solving capabilities from experience in a data-efficient manner. 
Several recent approaches \cite{ebert2018visual,zhang2018solar,Ha2018WorldModels,xie2019improvisation,nagabandi2019deep,hafner2019dream} employ powerful predictive models \cite{finn2016unsupervised, buesing2018learning, hafner2018learning,lee2018savp} %\todo{Cite non-vision work if necessary} 
to enable agents to predict and plan in complex environments directly from visual sensory observations, without needing to engineer a state estimator. % achieving sample-efficient control even from high-dimensional image observations. 
To plan a sequence of actions, these approaches usually use the predictive model to generate candidate roll-outs starting from the current state and then search for the sequence that best reaches the goal using a cost function (see Fig.~\ref{fig:teaser}, left). %A schematic visualization of this for a navigation task is shown in Fig.~\ref{fig:teaser}~(left).
However, such approaches do not scale to complex long-horizon tasks \cite{ebert2018visual}. %Oakland to San Francisco without a map.
%\FE{Are we putting the tree figure back in here? Should we replace Oakland with NYC?}
%%SL.5.28: without the old picture showing San Francisco planning, this sentence doesn't really make sense -- consider bringing that picture back?
Imagine the task of planning a route from your home to the airport. The above approaches would attempt to model all possible routes starting at home and then search for those that ended up at the airport. For long-horizon problems, the number of possible trajectories grows very large, making extensive search infeasible. 
%%DJ.6.1: Can we increase the font size in the image? This is way too small.

In contrast, we propose a planning agent that
only considers trajectories that start at home and end at the airport, i.e., makes predictions with the goal in mind. This approach both reduces prediction complexity as a simpler trajectory distribution needs to be modeled, and significantly reduces the search space for finding the best route, as depicted in Fig.~\ref{fig:teaser} (center). Indeed, we can produce a feasible plan simply as a single forward pass of the generative model, and can further refine it to find the optimal plan through iterative optimization.  %Using this approach, it is possible to plan a feasible trajectory independent of the horizon of the problem.  %since the generative model directly produces feasible trajectories, if the planning objective is to find a feasible trajectory, planning can be performed simply as a forward pass of the generative model, without the need for iterative optimization.

However, modeling this distribution becomes challenging for long time horizons even with goal-conditioned predictors. A naive method inspired by sequential predictive approaches would predict future trajectories at a fixed frequency, one step at a time --- the equivalent of starting to plan the route to the airport by predicting the very first footsteps. This can lead to large accumulating errors. Moreover, the optimization problem of finding the best trajectory remains challenging. The sequential planning approaches are unable to focus on large important decisions as most samples are spent optimizing local variation in the trajectory. To alleviate both shortcomings, %Instead, when faced with long-horizon tasks, we usually
we propose to predict an a tree-structured way, starting with a coarse trajectory and recursively filling in finer and finer details. This is achieved by recursive application of a single module that is trained to answer: given two states, what is a state that occurs between them? %intermediate goals, such as getting into the car or crossing the bridge. These intermediate goals hierarchically decompose the long-horizon task into shorter subtasks, as shown in Fig.~\ref{fig:teaser}~(right), allowing for more efficient prediction and planning.
This hierarchical prediction model is effective at long-term prediction and further enables us to design an efficient long-horizon planning approach by employing a coarse-to-fine trajectory optimization scheme. 
% \OVR{this paragraph seems a bit messy. It conflates two ideas: that hierarchy is important for prediction, and that hierarchical optimization is important for planning.}

Hierarchical prediction models naturally lend themselves to modeling the hierarchical structure present in many long-horizon tasks by breaking them into their constituent steps. However such procedural steps do not all occur on a regularly spaced schedule or last for equal lengths of time. Therefore, we further propose a version of our model based on a novel probabilistic formulation of dynamic time warping \cite{sakoe1971dynamic} that allows the model to select which frames to generate at each level in the tree, enabling flexible placement of intermediate predictions.

In summary, the contributions of this work are as follows. First, we propose a framework for goal-conditioned prediction and planning that is able to scale to visual observations by using a latent state model. Second, we extend this framework to hierarchical prediction and planning, which improves both efficiency and performance through the coarse-to-fine strategy and effective parallelization. We further extend this method to modeling the temporal variation in subtask structure. Evaluated on a complex visual navigation task, our method scales better than alternative approaches, allowing effective control on tasks longer than possible with prior visual planning methods.

\section{Related Work}
\label{scn:intro}
% \FE{If we don't get visual planning working we must remove the "visual" keyword from the following paragraph!}
% Here we discuss how our proposed GCP model relates to prior work on modeling dynamics and long-horizon planning.

%%SL.5.28: Remember to add citations that reviewers suggested last time if you haven't already!
% \textbf{Non-monotonic prediction}
% In many scenarios predicting in non-monotonic, non-sequential order can enable more accurate, long-term predictions. In \cite{tai2015improved, chung2016hierarchical, welleck2019non, liu2019naomi} a variable number of layers of temporal abstraction are used to model sequential data, showing improvements on prediction accuracy. We employ a similar model and show that it can be embedded it into a hierarchical planner, significantly outperform auto-regressive predictors.
%DJ.6.1: Add TD-VAE~\cite{gregor2018temporal} if space. Monotonic, but jumpy non-step-by-step predictions. Could name the paragraph temporally abstract prediction if including.

\textbf{Video interpolation.} We propose a latent variable goal-conditioned prediction model  that is able to handle high-dimensional image observations. This resembles prior work on video-interpolation where given a start and a goal image, images are filled in between. So far such work has focused on short-term interpolation, often using models based on optical flow \cite{liu2017video, jiang2018super, Niklaus_CVPR_2017, Niklaus_ICCV_2017}. Recent work has proposed neural network models that predict images directly, but this work still evaluates on short-horizon prediction \cite{wang2019point}. The models introduced in our work by contrast scale to video sequences of up to 500 time steps modelling complex distributions that exhibit multi-modality.
%DJ.6.1: This paragraph could be more aptly titled Video interpolation.

\textbf{Visual planning and control.} 
% Leveraging demonstrations is a fundamental component in many artificial agents capable of goal-driven long-horizon behavior. In imitation learning approaches \citep{sermanet2016unsupervised,edwards2018} the assumption is usually that visual data comes from an optimal or near-optimal policy. This assumption often holds for short-term problems and simplifies the learning problem substantially. However in many practical settings for long-horizon tasks, demonstrations become highly suboptimal and multi-modal, violating the assumptions of these methods. We show that sampling-based planning can be used even when using highly suboptimal and multi-model expert data. 
Most existing visual planning methods \citep{finn2017deep, paxton2018, xie2019improvisation, ebert2018visual, hafner2018learning} use model predictive control, computing plans forward in time by sampling state or action sequences. This quickly becomes computationally intractable for longer horizon problems, as the search complexity grows exponentially with the number of time steps \cite{ebert2018visual}. Instead, we propose a method for  goal-conditioned hierarchical planning, which is able to effectively scale to long horizons as it both reduces the search space and performs more efficient hierarchical optimization. \citet{ichter2018learnedltspace} %and \citet{rhinehart2019precog} 
also perform goal-conditioned planning by constraining the search space to trajectories that end at the goal, however, the method is only validated on low-dimensional states. %\citet{liu2019naomi} proposes a hierarchical method for low-dimensional sequence inpainting, but does not perform planning or control.
In this paper, % However, these methods are hard to scale to high-dimensional observations due to very multimodal trajectory distributions with such observations. In this paper,
we leverage latent state-space goal-conditioned predictors that scale to visual inputs and further improve the planning by using a hierarchical divide-and-conquer scheme. % and leverage a hierarchical model to increase search efficiency.
 %\OVR{I think somewhere here should also be a discussion of model-free methods and prior goal-conditioned work. Should we have a separate goal-conditioned control section?}
Other types of goal-conditioned control include inverse models and goal-conditioned imitative models \cite{pathak2018zero, yu2018one, torabi2018behavioral, smith2019}. However, these methods rely on imitation learning and are limited to settings where high-quality demonstrations are available. In contrast, our goal-conditioned planning and control method is able to optimize the trajectory it executes, and does not require optimal training data. %These are ill-suited for long horizons, since they are not designed to handle multi-modality and sub-optimality of the demonstrations.

\textbf{Hierarchical planning.} 
% explain hierchcial models with only one or two layers
%Hierarchical planning has long been considered by the AI community as the primary way to scale up planning methods for long horizons tasks, going back to the ABSTRIPS \cite{sacerdoti1974planning} and ALPINE \cite{knoblock1990learning} planners.  More recently, hierarchical planning methods were able to integrate physics-based approaches and scale to solving more complex tasks requiring motion planning \cite{kaelbling2010hierarchical}. However, these classical approaches use pre-defined symbolic descriptions of the environments, and moreover are often unable to form novel hierarchical abstractions, which limits their applicability to diverse real-world tasks. 
While hierarchical planning has been extensively explored in symbolic AI \cite{sacerdoti1974planning, knoblock1990learning, kaelbling2010hierarchical}, these approaches are unable to cope with raw (e.g., image-based) observations, limiting their ability to solve diverse real-world tasks.
Instead, we propose an approach that learns to perform hierarchical planning directly in terms of sensory observations, purely from data. Since our method does not require human-designed specification of tasks and environment, it is applicable in general settings where trajectory data can be collected.
Recently, a number of different hierarchical planning approaches have been introduced \cite{Jayaraman2018, pertsch2020keyin, fang2019dynamics, nair2019hierarchical, kim2019variational, nasiriany2019planning} that only work well with one or two layers of abstraction due to the architectural design or computational bottlenecks.
% explain other tree-strucuted hierarchical planners
Some of the few hierarchical planning approaches that have been shown to work with more than two layers of abstraction use tree-structured models \cite{chen2019tree, jurgenson2019sub, parascandolo2020divide}. However these models have not been shown to scale to high-dimensional spaces such as images. 
While also using a tree-structured model similar to our method, \citet{chen2019tree} make the assumption that the map in the physical workspace is known.
To the best of our knowledge, our proposed hierarchical planning algorithm is the first to use a variable number of abstraction layers while scaling to high-dimensional inputs such as images.

\section{Goal-Conditioned Prediction}
%%DJ.6.1: Candidate for vspace cutting.

In this section, we formalize the goal-condition prediction problem, and propose several models for goal-conditioned prediction, including both auto-regressive models and tree-structured models. To define the goal-conditioned prediction problem, consider a sequence of observations $[o_1, o_2, ... o_T]$ of length $T$. Standard forward prediction approaches (Fig~\ref{fig:gcp_pgm}, left) observe the first $k$ observations and synthesize the rest of the sequence. That is,
%they model $p(o_2, o_3, \dots o_{T-1}|o_1)$. 
they model $p(o_{k+1}, o_{k+2}, \dots o_{T-1}|o_1, o_2, \dots o_k)$. 
Instead, we would like our goal-conditioned predictors to produce intermediate observations given the first and last elements in the sequence (Fig~\ref{fig:gcp_pgm}, center and right). In other words, they must model $p(o_2, o_3, \dots o_{T-1}|o_1, o_T)$.
%DJ9.23: This needs to be explained in more detail and possibly moved into each of the forward prediction and recursive infilling sections separately. In each section, first explain: What are these states? What information should they express? e.g. We assume a compact latent state s in which the evolution of the video can be modeled as a Markov process, and such that each observation $o_t$ is purely a function of $s_t$. As such, $s_t$ should contain all the information from the history of the video that is relevant to predicting its future ...
%In order to build accurate and scalable predictive models, 
We propose several designs for goal-conditioned predictors that operate in learned compact state spaces for scalability and accuracy.
%We further propose different orders of $s_t$ prediction. 
% Fig~\ref{fig:gcp_pgm} shows schematics of these GCP designs, as well as of a standard forward predictor.   % We propose different options for how to factorize this joint distribution over the elements of the sequence.
%DJ9.23: Why does the state need to be "compact"?
%DJ9.23: I'd recommend not talking about "different orders" here. Makes no sense at this stage.
%%SL.5.28: For what it's worth, I liked the "San Francisco" planning figure -- it was very evocative, and I think if you have space, it would in the balance be better to include it if possible (e.g., in the intro)

\subsection{Goal-Conditioned Sequential Prediction}\label{sec:gcp-seq}

%%DJ9.23: Start by discussing what standard (non-goal-conditioned) forward predictive models do. Then discuss the SVG version, and then explain how goal-conditioning is simply introducing the additional o_T in the conditioning variables for the first predicted frame s_2. 
%%DJ9.23: In the standard forward prediction setting where only the starting frame (or starting K frames) are observed, it is natural to consider forward state-space models ... 

We first present a naive design for goal-conditioned prediction based on forward auto-regressive models. 
% In this model, we predict latent state representations sequentially in chronological order, from the start to the end, with the prediction at each point in time conditioned on the first and final observations as well as the previous latent state. 
% \cc{The resulting model can be factorized as follows:
% \begin{equation}
%       p(o_2, o_3, \dots o_{T-1}|o_1, o_T) = \prod_{t=2}^{T-1} p(o_t|o_{1:t-1}, o_T).
% \end{equation}
% }
%However, the approaches that model the observations directly in an autoregressive manner are known to scale poorly both in terms of computational efficiency and predictive performance \cite{denton2018stochastic,buesing2018learning,hafner2018learning}. Instead, we design a latent state-space model  (GCP-sequential, shown in Fig \ref{fig:gcp_pgm}, center)  that predicts in a latent space represented by a random variable $s_t$ and then decodes the observations with a decoder $p(o_t|s_t)$. The likelihood of this model factorizes as follows: 
%DJ.6.1: Omitted some stuff above that was confusing and/or unnecessary. Confirm that this is okay.
%OR.6.1: Seems fine, although our paper is in a big part about how to get the latent state part right. I am not sure what's the best thing to do here.
Autoregressive models operating directly on observations scale poorly in terms of computational efficiency and predictive performance~\cite{denton2018stochastic,buesing2018learning,hafner2018learning}. We design a latent state-space model  (GCP-sequential, shown in Fig \ref{fig:gcp_pgm}, center)  that predicts in a latent space represented by a random variable $s_t$ and then decodes the observations with a decoder $p(o_t|s_t)$. The latent state $s_t$ is used to allow handling partially observable settings. The likelihood of this model factorizes as follows: 
\begin{equation}
       p(o_2, o_3, \dots o_{T-1}|o_1, o_T) =
     \int p(o_2|s_2) p(s_2|o_1, o_{T}) 
     \prod_{t=3}^{T-1} p(o_t|s_t) p(s_t \vert s_{t-1}, o_1, o_T ) ds_{2:T-1}. 
\end{equation}
%We condition the first state on the start and goal, and condition all consecutive states on the previous state in addition to the start and goal observations. 
We show in Sec~\ref{sec:arch} that this model is simple to implement, and can build directly on previously proposed auto-regressive sequence prediction models. However, 
%While this model is simple to implement and can build directly on previously proposed auto-regressive video prediction models, as we will discuss in Section~\ref{sec:arch}, 
its computational complexity scales with the sequence length, as every state must be produced in sequence. As we show empirically, this approach also struggles with modeling longer sequences due to compounding errors, and is prone to ignoring the goal information on these longer sequences as very long-term dependencies have to be modeled when predicting the second observation from the first observation and the goal. % this approach does not account for the hierarchical structure of natural sequences, which contains events and sub-events, as depicted in Fig~\ref{fig:teaser}. %\KP{remove events, discuss compounding errors like in intro?}
%this particular way of conditioning on the goal observation arguably does not full account for the hierarchical process that underlies real-world planning, where higher-level subgoals may be produced first, and then the finer details are filled in conditioned on those subgoals.
%%DJ9.23: For inference? training?

%%DJ9.23: "A decoder produces the observation $o_t$ conditioned on each predicted state $s_t$."

\subsection{Goal-Conditioned Prediction by Recursive Infilling}\label{sec:gcp-tree}

\begin{figure*}
  \centering
  \includegraphics[width=\linewidth]{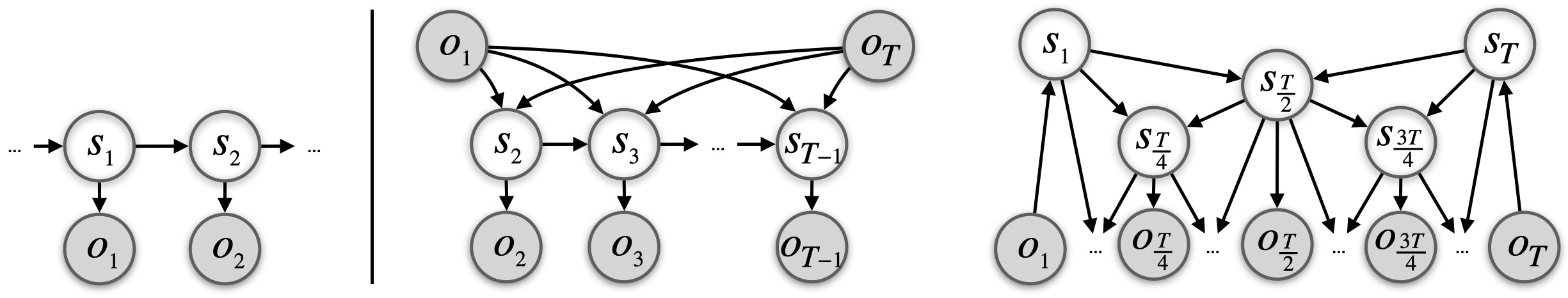}
%   \vspace{-0.3in}
  \caption{
   Graphical models for state-space sequence generation: forward prediction (left) and the proposed goal-conditioned predictors (GCPs). Shaded circles denote observations, white circles denote unobserved latent states. Center: a sequential goal-conditioned predictor with structure similar to forward prediction. Right: a hierarchical goal-conditioned predictor that recursively applies an infilling operator to generate the full sequence. All our models leverage stochastic latent states in order to handle complex high-dimensional observations. 
  }
%   \vspace{-0.1in}
  %DJ9.23: What is "video continuation"? Where is this referred to in text?
  %OR.6.1: addressed
  \label{fig:gcp_pgm}
\end{figure*}

%%DJ9.23: The discussion about what states are etc. should be somewhere here 

In order to scale goal-conditioned prediction to longer time horizons we now design a tree-structured GCP model that is both more efficient and more effective than the naive sequential predictor.
%GCP-sequential, described above, inherits the autoregressive forward propagation approach of standard forward prediction approaches. 
%The forward model only predicts based on the state at the previous timestep. 
%In the goal-conditioned setting, however, a future state is also available, and it is natural to condition predictions on both a past and a future state. 

Suppose that we have an intermediate state prediction operator $p(s_t|\text{pa}(t))$ that produces an intermediate latent state $s_t$ halfway in time between its two parent states $\text{pa}(t)$. Then, consider the following alternative process for goal-conditioned prediction depicted in Fig \ref{fig:gcp_pgm} (right): at the beginning, the observed first and last observation are encoded into the latent state space as $s_1$ and $s_T$, and the prediction operator $p(s_t|\text{pa}(t))$ generates $s_{T/2}$. 
%%DJ6.1: Not to be too nitpicky, but it's a bit annoying (to me at least!) that the numbering goes from 1 to T, and the middle frame is 0.5T not 0.5*(T+1). 
%%OR.6.1: Hmm I think these are actually the same if you round the first up and the second down. I would prefer to keep the T/2 notation since it's a bit cleaner and it fits in the figure. Addressed
The same operator may now be applied to two new sets of parents $(s_1, s_{T/2})$ and $(s_{T/2},s_T)$. As this process continues recursively, the intermediate prediction operator fills in more and more temporal detail until the full sequence is synthesized. %We can interpret filling in a state between parents as inferring an \textit{event} that must occur in between the parents. Thus we refer to this model as event-driven hierarchical model.%As we will discuss in Sec~\ref{sec:binding} in more detail, the intermediate frame mayeither be forced to bind to the middle frame each time, or may be allowed to select a frame to bind to.
%We will discuss the time index $t$ that this child index binds to, and the termination criterion for this recursive loop in Sec~\ref{Sec:binding}.

% Consider an intermediate state prediction operator $p(s_t|\text{pa}(t))$ that produces an intermediate state $s_t$ given its two parent states $\text{pa}(t)$. %How are these parent states assigned?
% At the beginning, the observed first and last frames are encoded into the latent state space as $s_1$ and $s_T$, and assigned to be the parents for the next prediction $s_t$.  The intermediate prediction operator $p(s_t|\text{pa}(t))$ generates $s_t$, which splits the full state sequence into two subsequences at time $t$. Each subsequence contains a start-goal pair, to which the same operator be applied. In this recursive process, the intermediate prediction operator fills in more and more temporal detail until the full video is synthesized. %As we will discuss in Sec~\ref{sec:binding} in more detail, the intermediate frame may either be forced to bind to the middle frame each time, or may be allowed to select a frame to bind to.
%We will discuss the time index $t$ that this child index binds to, and the termination criterion for this recursive loop in Sec~\ref{Sec:binding}.

%Doing this repeatedly results in prediction of the entire sequence by recursively filling its elements. 
We call this model GCP-tree, since it has a tree-like\footnote{The generation process closely mimics a graph-theoretic tree, but every node has two parents instead of one.} shape where each predicted state is dependent on its left and right parents, starting with the start and the goal. GCP-tree factorizes the goal-conditioned sequence generation problem as:
\begin{equation}
    \label{eq:gcp_tree}
     p(o_2, o_3, \dots o_{T-1}|o_1, o_T)  = \int p(s_1|o_1) p(s_T|o_T) 
     \prod_{t=2}^{T-1} p(o_t|s_t) p(s_t \vert \text{pa}(t)) ds_{1:T}.
\end{equation}
%%DJ.5.25: Shouldn't the integral here strictly be over ds1:T since you've put p(s_1|o_1) and p(s_T|o_T) in the integrand? 
%%OR.5.26: Indeed, fixed!
%%DJ.6.1: In that case, Fig 2c is a bit wrong. Might warrant fixing ... you could draw o_1 and o_T at the bottom to line up  with the other observations, and s_1 and s_T where o_1 and o_T currently are.
%%OR.6.1: addressed

%Here, $\text{pa}(n)$ represents the right and the left parents of the node associated with timestep $t$. 

%A simple choice for the intermediate frame prediction is to always produce a child state whose timestamp is exactly halfway between those of its two parents: $t_n = \ceil{({t_{pa_l(n)} + t_{pa_r(n)}})/{2}}$. 

% Such hierarchical prediction can enable more efficient prediction as it induces an independence structure that enables parallelization of the generation process.

\paragraph{Adaptive binding.}

%\OVR{This section is conditioned on whether we want to include it.}

%%SL.9.24: If you do include this, it's important to set up some motivation up front, e.g.: While the simplest way to formulate a tree-structured GCP model is to split the sequence precisely in half at each level of the tree, such balanced splits may not correspond to the natural bottlenecks in physical processes. For example, in the navigation example in Figure 1, we might prefer the first split to correspond to traversing the bridge, which partitions the prediction problem into two largely independent halves. To that end, we...

We have thus far described the intermediate prediction operator as always generating the state that occurs halfway in time between its two parents. While this is a simple and effective scheme, it may not correspond to the natural hierarchical structure in the sequence. For example, in the navigation example from the introduction, we might prefer the first split to correspond to visiting the bank, which partitions the prediction problem into two largely independent halves. We then design a version of GCP-tree that allows the intermediate state predictor to select which of the several states between the parents to predict, each time it is applied. In other words, the predicted state might \emph{bind} to one of many observations in the sequence.
%is capable of adaptive assignment of the child's node time step. 
In this more versatile model, we represent the time steps of the tree nodes with discrete latent variable $w$ that selects which nodes bind to which observations: $p(o_t|s_{1:N}, w_t) = p(o_t|s_{w_t})$. We can then express the prediction problem as:
\begin{multline*}
    p(o_{2:T-1}|o_1, o_T) = %\\
    \int  p(s_1|o_1) p(s_N|o_T) 
    \prod_{n} p(s_n \vert \text{pa}(n)) 
    \prod_{t=2}^{T-1} p(o_t|s_{1:N}, w_t) p(w_t) ds_{1:N} dw_{2:T-1}.
 \end{multline*}
%%DJ.5.25: The number of states N is still the same as the number of timesteps T, right? I see the reason for using a different index n rather than t for the states, but N instead of T is a bit confusing. 
%%OR.5.26: Here, it is actually not the same. We always generate the same number of states and the matching process figures out which ones are used in the loss.
Appendix \ref{seq:hedge} shows an efficient inference procedure for $w$ based on a novel probabilistic version of dynamic time warping \cite{sakoe1971dynamic}.

\subsection{Latent Variable Models for GCP}

% \begin{wrapfigure}{R}{0.5\textwidth}
%   \centering
%   \includegraphics[width=0.9\linewidth]{figures/gcp_latent_models.png}
%   \caption{
%   Proposed latent variable models. Circles represent stochastic variables, shaded circles represent variables observed at training and squares represent deterministic variables. \textbf{Left:} Sequential prediction. \textbf{Right:} Hierarchical prediction.
%   }
%   \vspace{-0.2in}
%   \label{fig:latent_pgm}
% \end{wrapfigure}

We have so far described the latent state $s_t$ as being a monolithic random variable. However, an appropriate design of $s_t$ is crucial for good performance: a purely deterministic $s_t$ might not be able to model the variation in the data, while a purely stochastic $s_t$ might lead to optimization challenges. Following prior work \cite{denton2018stochastic,hafner2018learning}, we therefore divide $s_t$ into $h_t$ and $z_t$, i.e. $s_t = (h_t, z_t)$, where $h_t$ is the deterministic memory state of a recurrent neural network, and $z_t$ is a stochastic per-time step latent variable. 
% To build powerful probabilistic models for goal-conditioned prediction, we propose to model the stochasticity with a per-node latent variable $z$ (see Fig \ref{fig:latent_pgm}), {inspired by prior work on forward prediction \cite{buesing2018learning, denton2018stochastic}}. 
To optimize the resulting model, we leverage amortized variational inference \cite{kingma2014auto,rezende2014stochastic} with an approximate posterior $q(\tilde{z} \vert o_{1:T})$, where $\tilde{z} = z_{2:T-1}$. The deterministic state $h_t$ does not require inference since it can simply be computed from the observed data $o_1, o_T$. The training objective is the following evidence lower bound on the log-likelihood of the sequence:
\begin{equation}
\label{eq:elbo}
    \ln p(o_{2:T-1} \vert o_{1,T})   \geq 
    \E_{q(\tilde z)} \left[\ln p(o_{2:T-1} \vert o_{1,T}, \tilde z)\right] -
     \text{KL}\left(q(\tilde z) \;\vert\vert\; p(\tilde z | o_{1,T})\right). 
    % \ln p(o_{2:T-1} \vert o_{1,T})   \geq \\
    % \E_{q(z_{2:T-1} \vert x)} \left[\ln p(o_{2:T-1} \vert o_{1,T}, z_{2:T-1})\right] -
    %  \text{KL}\left(q(z \vert o_{1:T}) \;\vert\vert\; p(z_{2:T-1} | o_{1,T})\right). 
\end{equation}
%%DJ.6.1: the first term in KL is missing subscripts 2:T-1 for z. I actually think something like \tilde{z} to represent z_{2:T-1} is fine notation for winning back space and getting the equation into one line. (if you introduce that in text).
%%OR.6.1: addressed
% In practice, we use a weight $\beta$ on the KL-divergence term, as is common in amortized variational inference~\citep{higgins2017beta, pmlr-v80-alemi18a, denton2018stochastic}. 
%\FE{we need to an equation to decompose the ELBO over time!}
%%DJ.6.1: I agree with Frederik above. I think expanding the terms in an appendix and pointing to that here would be good. Could draw a graph like Fig 2c but with h_t and z_t nodes.
%%OR.6.1: agreed, will add to appendix
%%DJ.6.1: Separately, do we cover the operation of the h_t recurrent network somewhere?
%%OR.6.1 the h_t should hopefully be clear from the architecture figure

\newpage

\subsection{Architectures for Goal-Conditioned Prediction}
\label{sec:arch}

\begin{wrapfigure}{R}{0.4\textwidth}
  \centering
  \vspace{-0.5in}
  \includegraphics[width=\linewidth]{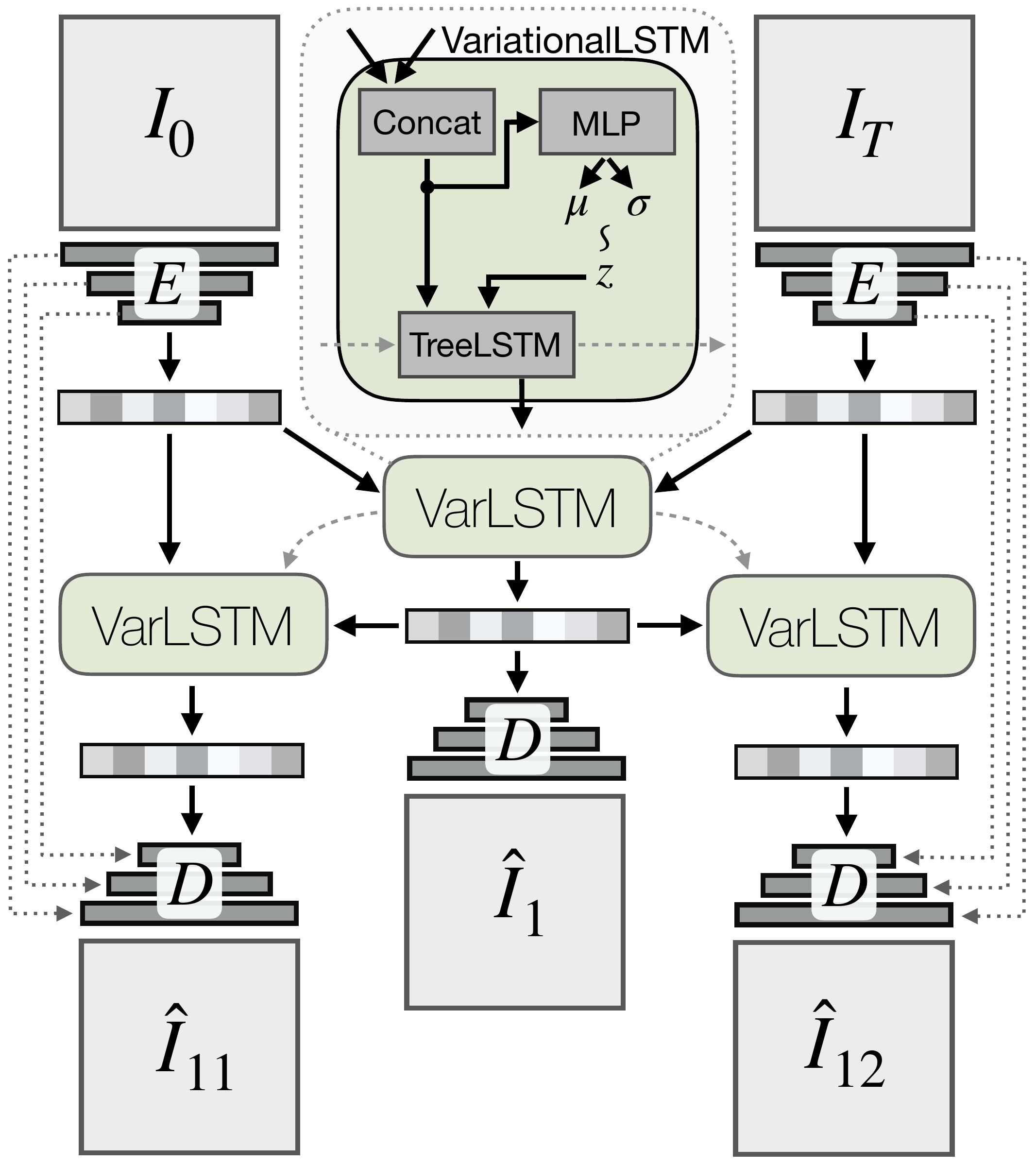}
  \vspace{-0.2in}
  \caption{
   Architecture for two-layer hierarchical goal-conditioned predictor (GCP). %and the first two layers of prediction. The VarLSTM module shows the computation of the latent state $s_t$, which consists of a deterministic and a stochastic part. See Section~\ref{sec:arch} for details. 
   Skip connections to first node's decoder omitted for clarity.
  } %%DJ.6.1: Nice figure, but could win some space back by making it more compact. The image boxes are taking lots of space. -- done
  %%DJ.6.1: What does VarLSTM stand for?
  \vspace{-0.2in}
  \label{fig:architecture}
\end{wrapfigure}

%%SL.9.24: There is some level of detail issue here. Can you first describe how this is a VAE-style model, etc., and only then dive into the details of how it's an LSTM network? Also make sure to actually reference the diagram figure!!

We describe how GCP models can be instantiated with deep neural networks to predict sequences of high-dimensional observations $o_{1:T}$, such as videos. %Concretely, the parts of the predictive model are implemented as follows: 
The prior $p(z_t| \text{pa}(t))$ is a diagonal Gaussian whose parameters are predicted with a multi-layer perceptron (MLP). The deterministic state predictor $p(h_t|z_t, \text{pa}(t))$ is implemented as an LSTM~\citep{hochreiter1997long}. %, and the decoding distribution $p(o_t|s_t)$. Additionally, we use an amortized variational inference network $q(z_t|o_{1:T})$.  
We condition the recurrent predictor on the start and goal observations encoded through a convolutional encoder $e_t = E(o_t)$. The decoding distribution $p(o_t|s_t)$ is predicted by a convolutional decoder with input features $\hat{e}_t$ and skip-connections from the encoder~\citep{villegas2017decomposing, denton2018stochastic}. In line with recent work \cite{rybkin2020sigmavae}, we found that learning a calibrated decoder is important for good performance, and we use the discrete logistics mixture as the decoding distribution \cite{salimans2017pixelcnn++}. The parameters of the diagonal Gaussian posterior distribution for each node, $q(z_t \vert o_{t}, \text{pa}(t))$, are predicted given the corresponding observation and parent nodes with another MLP. % via an attention mechanism over the embeddings of the evidence sequence \citep{bahdanau2014neural, luong2015effective}: $(\sigma_{q,t}, \mathcal{N}_{q,t}) = \text{MLP}_q(\text{Att}(E(o_{1:T}), \text{pa}(t))$. 
For a more detailed description of the architectural parameters we refer to Appendix~\ref{sec:app_architecture}.

\vspace{-0.05in}
\section{Planning \& Control with Goal-Conditioned Prediction}
\vspace{-0.051in}
\label{sec:planning_control}

In the previous section, we described an approach to goal-conditioned sequence prediction or GCP. The GCP model can be directly applied to control problems since, given a goal, it can produce realistic trajectories for reaching that goal. %In this section, we discuss the application of the GCP model for control, and propose an efficient hierarchical planning scheme that allows us to scale our GCP agent to challenging long-horizon tasks.
However, in many cases our objective is to reach the goal \emph{in a specific way}. For instance, we might want to spend the least amount of time or energy required to reach the goal. In those cases, explicit planning is required to obtain a trajectory from the model that optimizes a user-provided cost function $\mathcal{C}(o_t, \dots, o_{t^\prime})$. In GCPs, planning is performed over the latent variables $z$ that determine \emph{which} trajectory between start and goal is predicted: $\min_z \mathcal{C}(g(o_t, o_T, z))$, where $g$ is the GCP model. We propose to use the cross-entropy method (CEM, \cite{blossom2006cross}) for optimization, which has proven effective in prior work on visual MPC~\cite{ebert2018visual, nagabandi2019deep, nasiriany2019planning, pertsch2020keyin}. Once a trajectory is planned, we infer the actions necessary to execute it using a learned inverse model (see Appendix, Section~\ref{app:sec_planning_details}).

\begin{wrapfigure}{R}{0.55\textwidth}
\vspace{-0.35in}
\begin{minipage}{0.55\textwidth}
    \begin{algorithm}[H]
    \caption{Goal-Conditioned Hierarchical Planning}
    \label{alg:plan}
        \begin{algorithmic}[1]
            \STATE \textbf{Inputs:} Hierarchical goal-conditioned predictor $g$, current \& goal observation $o_t, o_T$, cost function $\hat{\mathcal{C}}$
            \STATE Initialize plan: $P = [o_t, o_T]$
            \FOR[iterate depth of hierarchy]{$d~=~1...D$}
            \FOR{$n~=~0...\vert P \vert-1$}
            \STATE $\mathbf{z}\sim \mathcal{N}(0, I)$ \COMMENT{sample M subgoal latents}
            \STATE $\mathbf{o}_{\text{sg}} = g(P[n], P[n+1], \mathbf{z})$ \COMMENT{predict subgoals}
            \STATE $o_{d,n} = \arg\min_{o \in \mathbf{o}_{\text{sg}}} \hat{\mathcal{C}}(P[n], o) + \hat{\mathcal{C}}(o, P[n+1])$
            \STATE \textsc{Insert}$(P, o_{d,n})$ \COMMENT{insert best subgoal in plan}
            \ENDFOR
            \ENDFOR
            \STATE \textbf{return }$P$
        \end{algorithmic}
        % \bottomrule
    \end{algorithm}
\end{minipage}
\vspace{-0.1in}
\end{wrapfigure}
\textbf{Goal-conditioned hierarchical planning.} 
Instead of optimizing the full trajectory at once, the hierarchical structure of the GCP-tree model allows us to design a more efficient, hierarchical planning scheme in which the trajectories between start and goal are optimized in a coarse-to-fine manner. % to reach the goal and minimize a user-provided cost function $\mathcal{C}(o_t, \dots, o_{t^\prime})$. 
The procedure is detailed in Algorithm~\ref{alg:plan}. We initialize the plan to consist of only start and goal observation. Then our approach recursively adds new subgoals to the plan, leading to a more and more detailed trajectory. Concretely, we proceed by optimizing the latent variables of the GCP-tree model $g(o_t, o_T, z)$ layer by layer: in every step we sample $M$ candidate latents per subgoal in the current layer and pick the corresponding subgoal that minimizes the total cost with respect to both its parents. The best subgoal gets inserted into the plan between its parents and the procedure recurses. 

\textbf{Cost function.} Evaluating the true cost function $\mathcal{C}(o_t, \dots, o_{t^\prime})$ would require unrolling the full prediction tree. For more efficient, \emph{hierarchical} planning, we instead want to evaluate the \emph{expected} cost $\hat{\mathcal{C}}(o_t, o_{t^\prime}) = \mathbb{E}_{(o_t, \dots, o_{t^\prime}) \sim \mathcal{D}} \;\mathcal{C}(o_t, \dots, o_{t^\prime})$ of a trajectory between two observations under the training data distribution $\mathcal{D}$. This allows us to estimate the cost of a trajectory passing through a given subgoal as the sum of the pairwise cost estimates to both its parents, without predicting all its children. 
We train a neural network estimator for the expected cost via supervised learning by randomly sampling two observations from a training trajectory and evaluating the true cost on the connecting trajectory segment $\mathcal{C}(o_t, \dots, o_{t^\prime})$ to obtain the target value.

\section{Experimental Evaluation}

The aim of our experiments is to study the following questions: (1) Are the proposed GCPs able to effectively predict goal-directed trajectories in the image space and scale to long time horizons? (2) Is the proposed goal-conditioned hierarchical planning method able to solve long-horizon visual control tasks? %2) Does the event-driven hierarchical model find high-level events in the trajectories
% 2) Does the proposed method for goal-conditioned hierarchical prediction scale to modeling long-horizon sequences of high-dimensional data such as video and can the hierarchical structure enable faster training on such long sequences?
(3) Does the version of GCP with adaptive binding find high-level events in the trajectories?
%%CF.2.3: right now there's a subsection on runtime, but not a question about runtime. would be good to add one, or at least allude to it in one of the above.
%2) Does tree-structured prediction improve efficiency for long-horizon video prediction? 3) Does goal-conditioned prediction enable long-term imitation without access to the expert's actions?

% \paragraph{Datasets.}

% \begin{wrapfigure}{R}{0.5\textwidth}
%   \centering
%   \vspace{-5pt}
%   \includegraphics[width=\linewidth]{figures/gcp_datasets.png}
%   \vspace{-10pt}
%   \caption{Datasets used for our evaluation of goal-conditioned prediction. \textbf{Left to right}: {Human 3.6M} (64$\times$64px), \pp dataset (64$\times$64px) and Maze dataset (16$\times$16px)
%   }
% %\vspace{-20pt}
%   \label{fig:dataset_overview}
% \end{wrapfigure}

\subsection{Goal-Conditioned Video Prediction}
\label{sec:video_pred}

\begin{table*}[t]
\caption{%
Long-term prediction performance of the goal-conditioned predictors compared to prior work on video interpolation. 
Additional evaluation on FVD / LPIPS~\cite{unterthiner2018towards,zhang2018perceptual} in Appendix, Table~\ref{tab:app_perceptual_metrics}.
}
\begin{center}
\begin{footnotesize}
\begin{sc}
\begin{tabular}{lllllllll}
\toprule
\footnotesize
Dataset  &  \multicolumn{2}{c}{Pick\&Place} &  \multicolumn{2}{c}{Human 3.6M} & \multicolumn{2}{c}{9 rooms Nav} & \multicolumn{2}{c}{25 rooms Nav}\\
\cmidrule(r){2-3} \cmidrule(r){4-5} \cmidrule(r){6-7} \cmidrule(r){8-9} 
Method & PSNR & SSIM & PSNR & SSIM & PSNR & SSIM & PSNR & SSIM\\% & PSNR-DTW & SSIM-DTW\\
\midrule

GCP-tree 
& \textbf{34.34} & \textbf{0.965}
& \textbf{28.34} & \textbf{0.928} 
& \textbf{13.83} & \textbf{0.288}  

& \textbf{12.88} & \textbf{0.279}  \\

GCP-sequential 
& \textbf{34.45} & \textbf{0.965}
& 27.57 & 0.924
& {12.91} & 0.213
& {11.61} & {0.209}  \\

\midrule
DVF \cite{liu2017video}
& 26.15 & 0.858  
& 26.74 & 0.922
& 11.678 & 0.22
& 11.34 & 0.172 \\  

CIGAN \cite{kurutach2018learning}
& 21.16 & 0.613
& 16.89 & 0.453
& 11.96 & 0.222
& 9.91 & 0.150  \\

\bottomrule
\end{tabular}
\end{sc}
\end{footnotesize}
\end{center}
\label{tab:vid_pred}
\vskip -0.2in
\end{table*}
\begin{wraptable}{R}{0.4\linewidth}
\vskip -0.25in
\caption{Ablation of prediction performance on \pp} \label{tab:ablation}
\begin{center}
\begin{sc}
\vskip -0.15in
\begin{tabular}{lcccc}
\toprule
Method & PSNR & SSIM \\
\midrule
Tree   & $34.34$  &  0.965  \\
Tree w/o skips   & $32.64$  &  0.955  \\
Tree w/o LSTM  & $31.44$  &  0.947  \\

\bottomrule
\end{tabular}
\end{sc}
\end{center}
\vskip -0.2in
\end{wraptable}

%%SL.2.2: need a transition sentence to explain why you're suddenly talking about this (i.e., In this section, we focus on the predictive performance of GCP models, on visually more complex datasets...)

Most commonly used video datasets in the literature depict relatively short motions, making them poorly suited for studying long-horizon prediction capability. We therefore evaluate on one standard dataset, and two synthetic datasets that we designed specifically for evaluating long-horizon prediction. %We use the original frequency of 50Hz.
% To evaluate on real-world videos, we use the {Human 3.6M} dataset \citep{ionescu2013human3}, which depicts multiple different people performing various tasks. %, and a simulated dataset of robotic pushing interactions from \citep{Jayaraman2018}, which contains video sequences of a robot arm pushing three objects one-by-one.
The \pp dataset contains videos of a simulated Sawyer robot arm placing objects into a bin. Training trajectories contain up to 80 frames at $64\times64$~px and are collected using a simple rule-based policy. The \textit{Navigation} data consists of videos of an agent navigating a simulated environment %based on the Gym-Miniworld \citep{gym_miniworld}. %, which constists of a fixed grid room layout.
with multiple rooms:
we evaluate versions with 9-room and 25-room layouts, both of which use $32\times32$~px agent-centric top-down image observations, with up to 100 and 200 frame sequences, respectively. We collect example trajectories that reach goals in a randomized, suboptimal manner, providing a very diverse set of trajectories (details are in App.~\ref{sec:app_data_maze}). %\footnote{Our model can even learn to plan using data collected with completely random actions (see supp. section~\ref{sec:random_data})}.
%%SL.5.28: Maybe provide a little bit more detail about how these trajectories are noisy and randomized? % which consists of a number of sparsely connected rooms located on a grid. The maze layout is constructed such that one single path exists between every pair of rooms and the wall texture of each room is unique. Demonstrations between random start and goal positions are collected using the probabilistic roadmap (PRM) planner~\citep{kavraki1996probabilistic}, leading to demonstrations with substantial noise.
% We evaluate the prediction performance of our method on sequences collected in a 3$\times$3 room and a 10$\times$10 room layout.
%%SL.5.28: not clear what 3x3 and 10x10 refers to, could phrase as: We use two versions of this dataset, one depicting navigation of a maze with 9 rooms, arranged in a 3$\times$3 pattern, and the other with 100 rooms, in a 10$\times$10 layout. We show overhead images of the 10$\times$10 maze in Figure ??.
We further evaluate on the real-world Human 3.6M video dataset \citep{ionescu2013human3}, %a dataset of real-world videos depicting multiple different people performing various tasks. We evaluate on
%in the challenging setting of
predicting $64\times64$~px frames at full frequency of 50Hz up to 10 seconds in the future to show the scalability of our method. This is in contrast to prior work which %subsampled the videos, %we evaluate on the original frequency of 50Hz to show the scalability of our method. We note that these sequences are substantially longer than the sequences that were used in prior work to evaluate video prediction models, which are
evaluated on subsampled sequences shorter than 100 frames (see \cite{denton2017unsupervised,denton2018stochastic,wichers2018hierarchical}). Architecture and hyperparameters are detailed in Appendix~\ref{sec:app_architecture}.
% We use 64$\times$64px spatial resolution for {Human 3.6M} and \pp and train on sequences of 500 frames and 80 frames respectively. For the Maze, we generate sequences up to 100 frames in length on the 3$\times$3 layout and 1000 frames for the 10$\times$10 rooms. Due to the very long sequences we evaluate on a resolution of 16$\times$16px.
%%SL.5.28: Maybe add something like: Note that these sequences are substantially longer than those typically used in evaluations of video prediction models. [maybe also include some citations and specific examples?]

%For the 25-room navigation data, we use a separate set of weights for each layer of hierarchy to increase the capacity of the GCP-tree model, and correspondingly increase the capacity of the other models. \todo{where are we describing the hyperparameters?}
%%KP.6.1: move this to appendix

\begin{wraptable}{R}{0.4\linewidth}
\vskip -0.25in
\caption{GCP runtime on $16\times16$~px H3.6M sequences in sec/training batch\protect\footnotemark } \label{tab:runtime}
\begin{center}
\begin{small}
\begin{sc}
% \vskip -0.05in
% \resizebox{0.4\textwidth}{!}{
\begin{tabular}{lccc}
\toprule
Seq Length & 100 & 500 & 1000  \\
\midrule
GCP-Seq & $1.49$ & $8.44$ & $17.6$\\
GCP-Tree & $\textbf{0.55}$ & $\textbf{1.66}$ & $\textbf{2.77}$\\
\midrule
Speed-up &  $\times2.7$ & $\times5.1$ & $\times6.4$\\
\bottomrule
\end{tabular}
% }
\end{sc}
\end{small}
\end{center}
\vskip -0.2in
%\end{wraptable}
\end{wraptable}
\footnotetext{We use $16\times16$~px to fit the 1000-frame sequences on a single NVIDIA V100 GPU, we expect the results to translate to larger resolutions on GPUs with larger memory.}

\begin{figure}
  \centering
  \includegraphics[width=\linewidth]{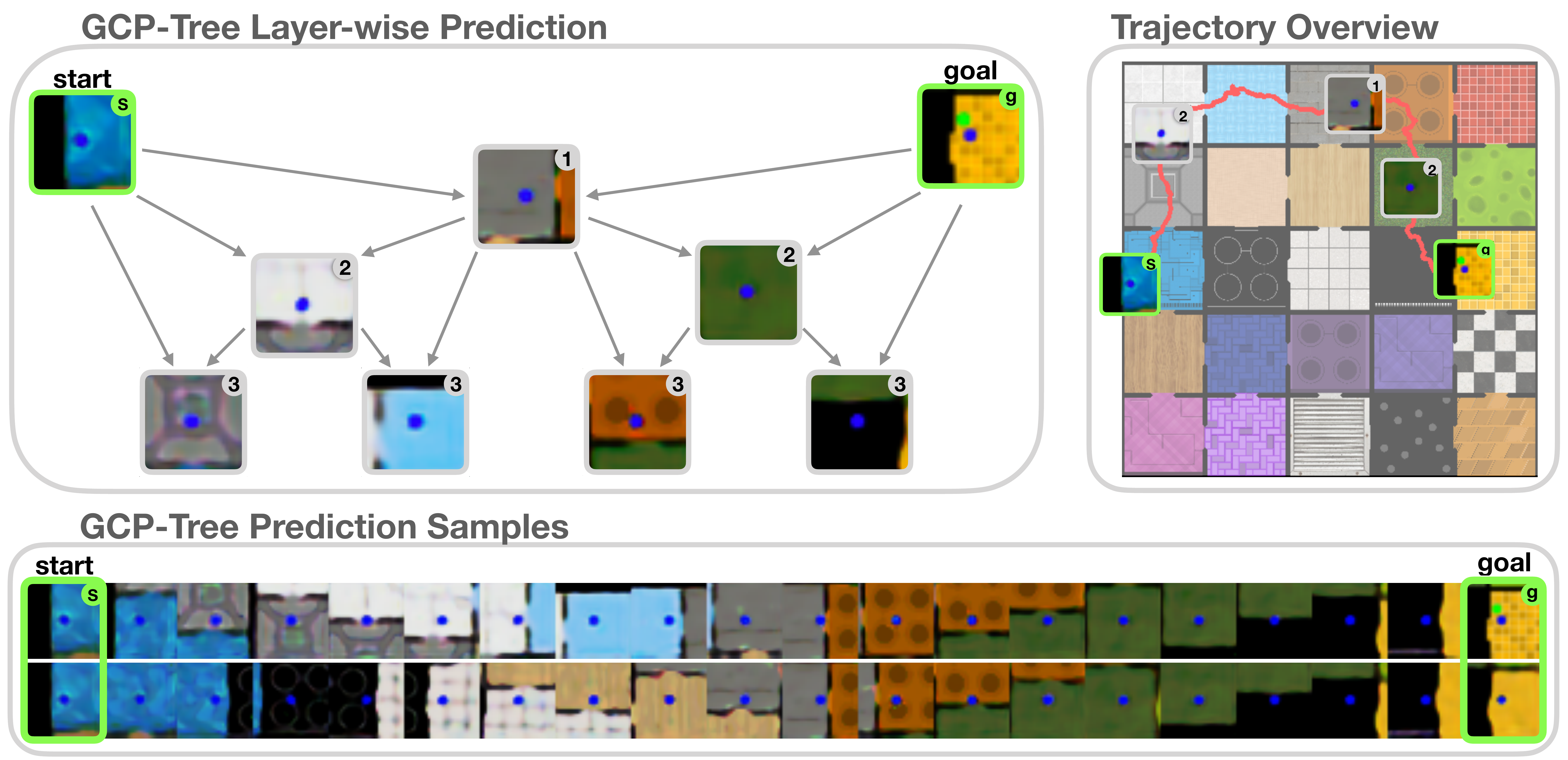}
  \vspace{-0.15in}
  \caption{Samples from GCP-tree on the 25-room data. \textbf{Left}: hierarchical prediction process. At each layer, the infilling operator is applied between every two frames, producing a sequence with a finer and finer temporal resolution. Three layers out of eight are shown. \textbf{Right}: visualization of the trajectory on the map together with a plan execution (see Section~\ref{sec:control_exp}).  \textbf{Bottom}: two image sequences sampled given the same start and goal (subsampled to 20 frames for visualization). Our model leverages stochastic latent states that enable modeling multimodal trajectories.
  See additional video results on the supplementary website: \url{orybkin.github.io/video-gcp}.}
  \vspace{-0.2in}
  \label{fig:predictions}
\end{figure}

In Tab.~\ref{tab:vid_pred}, we compare the GCP models to a state-of-the-art deep video interpolation method, DVF \citep{liu2017video},\footnote{While DVF has an official trained model, we re-train DVF on each dataset for better performance.} as well as a method for goal-conditioned generation of visual plans by interpolation in a learned latent space, CIGAN \cite{kurutach2018learning}. Following the standard procedure for evaluation of stochastic prediction models, we report top-of-100 peak signal-to-noise ratio (PSNR) and structural similarity metric (SSIM). %to the state-of-the art stochastic predictive model for temporal sequence data \citep{denton2018stochastic} that uses the standard forward prediction scheme: given a start image, a stochastic RNN predicts the image at the next timestep. This prior method uses autoregressive prediction by feeding the predicted image from the last timestep as input for the next prediction step. We condition this baseline on the goal frame at every timestep to produce goal-conditional predictions. It uses the same decoder architecture described in Sec.~\ref{sec:architecture}.
%%DJ.6.1: Could you put the citations for DVG, CIGAN, SVG into the table?
%%Done
%%DJ.6.1: Didn't we evaluate against Wichers et al for a rebuttal? Could we add it in here?
%%OR.6.1: that comparison didn't work out great because of technical issues...
%We observe that the proposed goal-conditioned prediction models outperform both interpolation baselines by a large margin.
We observe that the interpolation methods fail to learn meaningful long-term dynamics, and instead blend between start and goal image or predict physically implausible changes in the scene. %(see qualitative results in appendix).
%%SL.5.28: Say which appendix
%CIGAN, which uses latent space interpolation, similarly struggles to capture long-term transformations, predicting physically implausible changes in the scene.
In contrast, GCP-sequential and GCP-tree, equipped with powerful latent variable models, learn to predict rich scene dynamics between distant start and goal frames %, synthesizing sequences that traverse tens of different rooms
(see qualitative results in Fig.~\ref{fig:predictions} and for all methods on the project website. %\footnote{See additional video results on the supplementary website \url{orybkin.github.io/video-gcp}}).
%%SL.5.28: I think the reader will get pretty suspicious here that you are hiding the results in the appendix and not showing them here. Not certain what best to do about this, but keep in mind that it is likely to be an issue.
%%OR.6.1: perhaps a website is actually less suspicious here
% We attribute these results to the more powerful stochastic latent variable model our methods employ.

On the longer Human 3.6M and 25-room datasets, the GCP-tree model significantly outperforms the GCP-sequential model. Qualitatively, we observe that the sequential model struggles to take into account the goal information on the longer sequences, as this requires modeling long-term dependencies, while the hierarchical model is able to naturally incorporate the goal information in the recursive infilling process. Additionally, the hierarchical structure of GCP-tree enables substantially faster runtimes (see Table~\ref{tab:runtime}). We present an ablation study for GCP-tree in Tab.~\ref{tab:ablation}, showing that both the skip connections and the recurrence in the predictive module contribute to good performance.%Our proposed hierarchical approach, however, is able to predict well even on these more challenging datasets.    \todo{add evidence}

\subsection{Visual Goal-Conditioned Planning and Control}

\label{sec:control_exp}

\begin{wraptable}{R}{0.65\linewidth}
\vskip -0.25in
\caption{Visual control performance on navigation tasks} \label{tab:control}
\begin{center}
\begin{small}
\begin{sc}
\vskip -0.15in
\resizebox{0.65\textwidth}{!}{
    \begin{tabular}{lccccc}
    \toprule
    Method & \multicolumn{2}{c}{9-room Nav} & \multicolumn{2}{c}{25-room Nav}  \\
    \cmidrule(r){2-3} \cmidrule(r){4-5}
           & Success & Cost & Success & Cost \\
    \midrule
    GC BC \cite{nair2017combining} & $45\%$ & $139.75$ & $7\%$ & 402.48 \\
    VF \cite{ebert2018visual} & $84\%$ & $128.00$ & $26\%$ & 362.82\\
    Ours &  $\mathbf{93}\%$ & $\mathbf{34.34}$ & $\mathbf{82}\%$ & $\mathbf{158.06}$\\
    \midrule 
    GCP-flat & $\mathbf{94}\%$ & $36.00$ & $79\%$ & 181.02\\
    GCP-sequential & $91\%$ & $50.02$ & $14\%$ & 391.99\\
    \bottomrule
    \end{tabular}
}
\end{sc}
\end{small}
\end{center}
\vskip -0.2in
\end{wraptable}

% GCPs can be utilized for control in order to reach user-specified goal observations, e.g. an image of the desired outcome. In our prototype, we train GCPs on data that consists of trajectories that reach a wide variety of different goals, which can be obtained, for example, from an expert agent. Note that, in contrast to conventional imitation learning \cite{billard2008survey,hussein2017imitation}, GCP-based planning does not require demonstrations of reaching the test-time goal but instead can be used to reach \emph{new} goals, from within the distribution of training goals.

% \input{floats/state_prediction}

% We evaluate whether goal-conditioned visual prediction is suitable for imitating long-term, goal-directed expert behavior without access to the expert's actions. We first collect a dataset of 30k expert trajectories with up to 100 frames in the 3$\times$3 Maze environment. For planning and execution we follow the procedure detailed in Sec.~\ref{sec:imitation}. We train the inverse model that is used to follow the visual plans using 20k trajectories of 15 frames each, collected by taking random actions from random initial positions in the maze.

Next, we evaluate our hierarchical goal-conditioned planning approach (see Section~\ref{sec:planning_control}) on long-horizon visual control tasks. We test our method on a challenging image-based navigation task in the 9 and 25-rooms environments described in Section~\ref{sec:video_pred}. %Given a top-down image around the agent's current location as well as a goal image,
We use the data from Section~\ref{sec:video_pred} to train the predictive model. We note that our method does not require optimal demonstrations, but only data that is sufficiently diverse. Such dataset might be collected e.g. via crowd-sourced teleoperation \cite{zhu2020robosuite}, or with a suitable exploration policy \cite{eysenbach2018diversity,sekar2020planning}. For evaluation with purely random data, see supp. section \ref{sec:random_data}.
Given the current image observation the agent is tasked to reach the goal, defined by a goal image, on the shortest possible path.  %while minimizing a cost. We set the cost function $\mathcal{C}(o_t, \dots, o_{t^\prime})$ to be equal to the number of steps $t^\prime - t$ in the trajectory. %To more realistically model the local state representation typically available in navigation agents all observations are cropped to a small region around the agent (see Fig.~\ref{fig:plan}).
%To evaluate the influence of task horizon on performance, w
We average performance over 100 task instances for evaluation. These tasks involve crossing up to three and up to 10 rooms respectively, requiring planning over horizons of several hundred time steps, much longer than in previous visual planning methods~\cite{ebert2017self, ebert2018visual}.
We compare hierarchical planning with GCP to visual foresight (VF, \citet{ebert2018visual}), which optimizes rollouts from an action-conditioned forward prediction model via CEM~\cite{blossom2006cross}. We adopt the improvements to the sampling and CEM procedure introduced in \citet{nagabandi2019deep}. %and Causal InfoGan (CIGAN, \cite{kurutach2018learning}).
We also compare to goal-conditioned behavioral cloning~(GC BC, \cite{nair2017combining}% \todo{this paper uses HER+GAIL, which is completely not what our baseline is}
) as a ``planning-free" approach for learning goal-reaching from example goal-reaching behavior. %: we train a model that maximizes the likelihood of the training action trajectories conditioned on current observation and goal.
%%DJ.6.1: CITE THE MOST CLOSELY RELATED GOAL-CONDITIONED BC WORK TO THIS BASELINE. If significantly different from prior work, is there somewhere else where this is described in more detail? -- done
%%SL.2.2: Once we adjust Sec 4 to properly describe out method *not* as an imitation learning approach but rather as a method that plans with predictive models, it will no longer be obvious to the reader why the comparison with behavioral cloning makes sense, or even how to do it. Maybe make it clear that we are talking about goal-conditioned behavioral cloning (reference eg GCSL paper or Lynch et al), and briefly explain how it works?
%We first check the prediction quality of our method and ablate the key architectural components in Tab. \ref{tab:state_pred}. We see that conditioning the prediction on the goal substantially improves prediction accuracy and, moreover, the hierarchical prediction scheme significantly improves over the performance of the non-hierarchical model. %This is likely because the hierarchical model is better structured for long-term prediction, as the information can travel easily through the higher levels, which act as temporal skip-connections.

In Table~\ref{tab:control}, we report the average success rate of reaching the goal room, as well as the average cost, which corresponds to the trajectory length.\footnote{Since reporting length for failed cases would skew the results towards methods that produce short, unsuccessful trajectories, we report a constant large length for failed trajectories.} VF performs well on the easy task set, which requires planning horizons similar to prior work on VF, but struggles on the longer tasks as the search space becomes large. %Planning over such long horizons with CEM and forward prediction is difficult, and it is known that long horizons present a challenge for such model-based RL methods~\citep{janner2019trust}.
The BC method is not able to model the complexity of the training data and fails to solve these environments. In contrast, our approach %is able to leverage goal-conditioned hierarchical prediction to achieve the best planning performance, scaling
performs well %beyond time horizons possible with prior visual planning approaches.
even on the long-horizon task set.
% what are good other references for the model claim above?

%We show a qualitative example plan of our hierarchical planning approach for a long-horizon goal in Fig.~\ref{fig:plan},
%%DJ.6.1: Check for consistency in how figures, tables, sections etc. are referenced in text Fig vs Fig. vs Figure etc.
%Figure~\ref{fig:plan} shows an example plan generated by our hierarchical planning approach; further examples are in Appendix, Figure~\ref{fig:qual_hplan}.
%We ablate the effects of different components of our planning approach in Table~\ref{tab:control}. We find that when using the GCP model with a non-hierarchical planning scheme similar to \cite{ebert2018visual, nagabandi2019deep} the agent is still able to reach the goal room most of the time, however the executed trajectories are on average substantially longer than those optimized with the hierarchical planning approach. Using the GCP-sequential model instead of GCP-tree for sampling performs well on short tasks, but struggles to scale to longer tasks, highlighting the importance of the powerful hierarchical prediction model. %The ablation that uses the non-hierarchical GCP-sequential model for planning performs well on short tasks, but struggles to solve long-horizon tasks, which we attribute to the challenges of modeling long-range temporal dependencies to faraway goals in a non-hierarchical architecture.

\begin{figure}
  \centering
  \includegraphics[width=\linewidth]{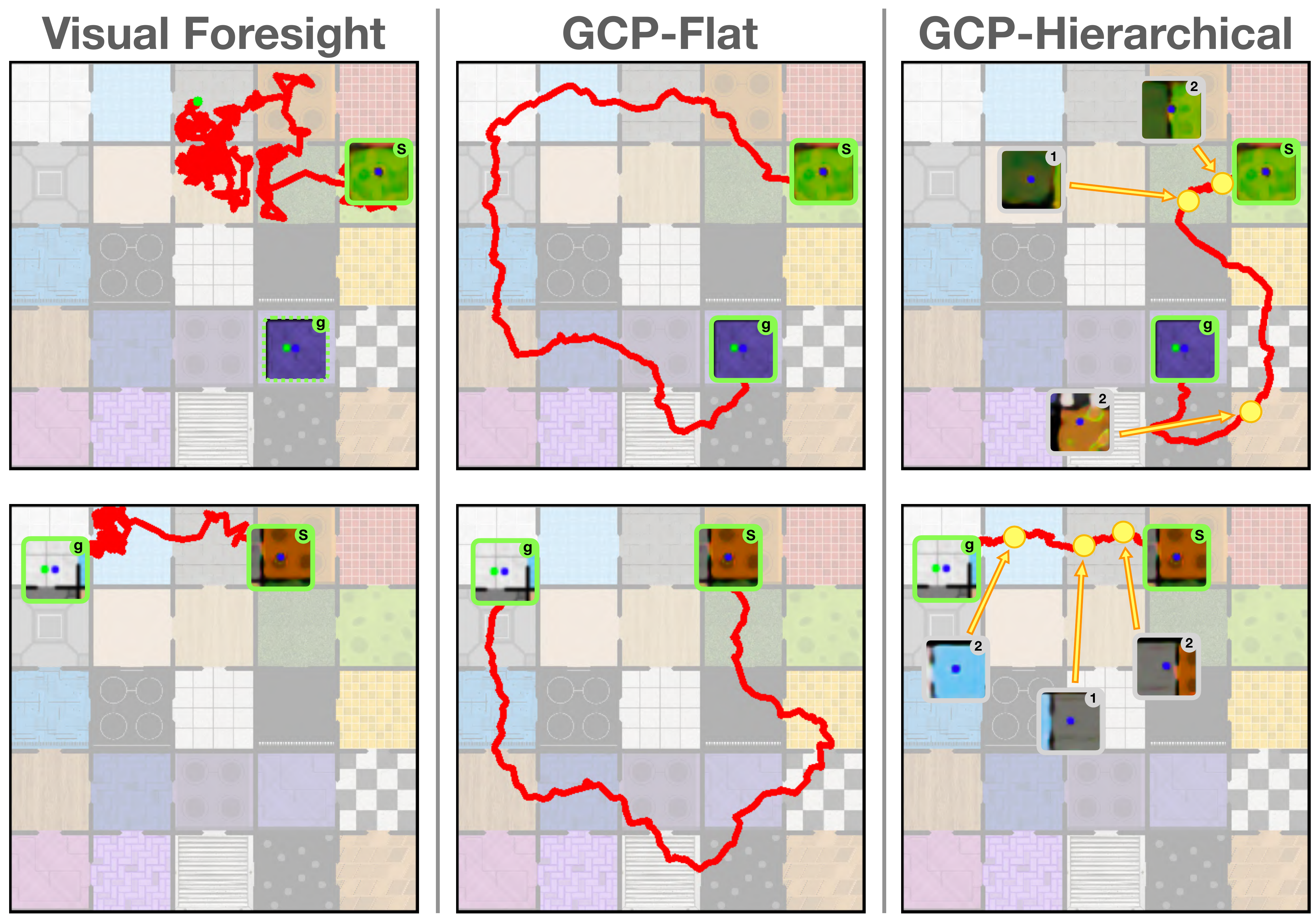}
   \vskip -0.1in
  \caption{Comparison between planning methods. Trajectories (red) sampled while planning from start (blue) to goal (green). All methods predict image trajectories, which are shown as 2d states for visualization. \textbf{Left}: visual MPC \cite{ebert2018visual} with forward predictor, \textbf{middle}: non-hierarchical planning with goal-conditioned predictor (GCP), \textbf{right}: hierarchical planning with GCP (ours) recursively optimizes subgoals (yellow/red) in a coarse-to-fine manner and finally plans short trajectories between the subgoals. Goal-conditioning ensures that trajectories reach the long-horizon goal, while hierarchical planning decomposes the task into shorter segments which are easier to optimize.% during planning.
  }
   \vskip -0.15in
%   \footnotetext{Trajectories were obtained by training a probe network that infers 2D coordinates from predicted images for easier visualization.}
  \label{fig:plan_progress}
\end{figure}
%%DJ.6.1. Would be cool to see a video of this plan being generated over time!

\begin{wrapfigure}{R}{0.6\textwidth}
  \centering
  \vskip -0.1in
  \includegraphics[width=\linewidth]{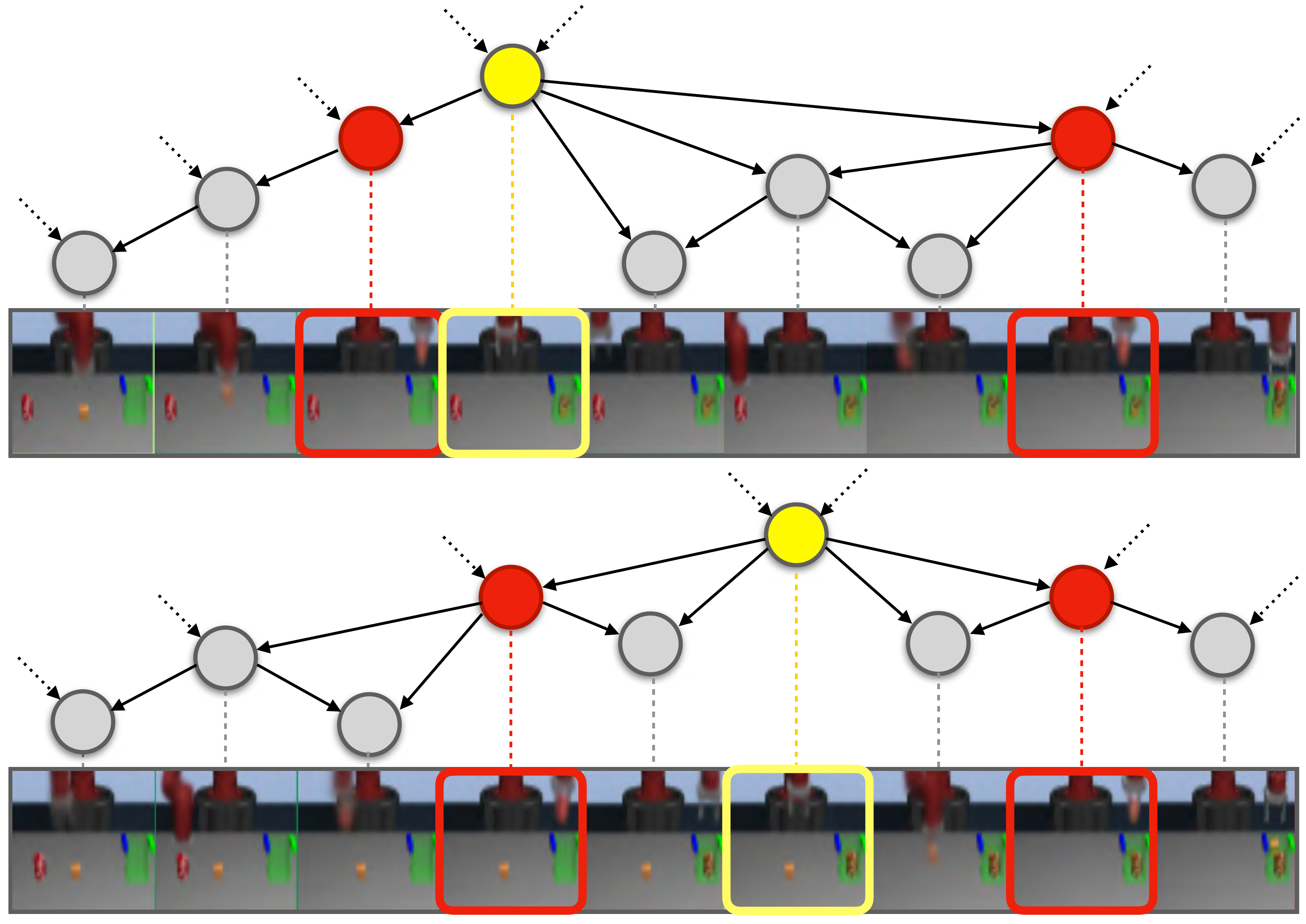}
  \vskip -0.05in
  \caption{
   Temporal abstraction discovery on \pp. Discovered tree structure with adaptive binding: nodes from the first two layers (yellow/red) bind to semantically consistent bottlenecks across sequences, e.g. in which the robot is about to drop the object into the bin.
   %Four predicted sequences for the adaptive GCP-tree model are shown, with a node in the second layer of the tree marked red.
   %%SL.5.28: I don't think it's clear what "a node in the second layer of the tree" means -- maybe just say: with the bottleneck discovered at the second layer of the tree marked in red. [that said, I think a visualization more like what is in Figure 6 would be more effective!]
   %This node reliably identifies and predicts a semantic bottleneck in the trajectory, corresponding to the frame where the robot is about to drop the object into the bin.
  }
  \label{fig:bottleneck}
  \vskip -0.2in
\end{wrapfigure}

We compare different planning approaches in Fig.~\ref{fig:plan_progress}. We find that samples from the forward prediction model in VF have low probability of reaching long-horizon goals. Using GCPs with a non-hierarchical planning scheme similar to \cite{ebert2018visual, nagabandi2019deep} (GCP-Flat) requires optimization over a large set of possible trajectories between start and goal and can struggle to find a plan with low cost. In contrast, our hierarchical planning approach finds plans with low cost by breaking the long-horizon task into shorter subtasks through multiple recursions of subgoal planning. % and then effectively optimizes the shorter intervals between subgoals, resulting in plans with overall low cost.
Using GCP-sequential instead of GCP-tree for sampling performs well on short tasks, but struggles to scale to longer tasks (see Table~\ref{tab:control}), highlighting the importance of the hierarchical prediction model.

\subsection{Temporal Abstraction Discovery}

%%DJ.6.1. Could we go with something more like Temporal Abstraction Discovery? I don't feel strongly about it, just a suggestion.
% \vskip -0.1in
We qualitatively evaluate the ability of GCP-tree \emph{with adaptive binding} (see Section~\ref{sec:gcp-tree}) to learn the temporal structure in the robotic \pp dataset. %, which consists of trajectories of a robotic arm picking and placing multiple objects into a bin. %decoder distribution variance
%%DJ.6.1: Point to where in the approach this model was described, since the reader may have long-forgotten this. Wouldn't hurt to also remind them of the nice DTW reformulation in appendix.
We increase the reconstruction loss of the nodes in the first two layers of the tree 50 times, forcing these nodes to bind to the frames for which the prediction is the most confident, the bottlenecks (see experimental details in Appendix~\ref{seq:hedge}).
%%SL.5.28: Introduce briefly the dataset that you are doing this with, so that it's obvious why these are the bottlenecks
%%KP.6.1: we introduce the dataset in Sec5.1 and describe the bottleneck below

In Fig. \ref{fig:bottleneck}, we see that this structural prior causes the model to bind the top nodes to frames that represent semantic bottlenecks, e.g. when the robot is about to drop the object in the bin. We found that all three top layer nodes specialize on binding to distinctive bottlenecks, leading to diverse predicted tree structures.
%%SL.5.28: Similar as in they look the same, or they are also bottlenecks (but other bottlenecks)?
We did not observe that adaptive binding improves the quality of predictions on our datasets, though the ability to discover meaningful bottlenecks may itself be useful~\citep{hrl,kipf2018compositional,infobot}. %, as discussed in the literature on hierarchical RL~\citep{hrl,kipf2018compositional} and exploration~\citep{infobot}.

\section{Discussion} % and Future Work}
\label{scn:conc}

%We presented a goal-conditioned predictor model (GCP) and a long-horizon planning framework based on this model. The GCP model generates plausible state sequences between a given start and goal state, learning the mechanics of the environment from data. %GCPs must learn to understand the mechanics of the environment that they are trained in, in order to accurately predict the intermediate events that must take place in order to bring about the goal images from the start images.
%%CF.2.3: the above sentence is really hard to follow.
%We explore how we can devise hierarchical GCP models that are suited for efficient hierarchical planning and suffer less from error accumulation when performing long-term prediction.
%%SL.5.28: Should say something about tree-structured model above! E.g., We present both a standard sequential variant of our model, and a hierarchical tree-structured variant, where the latter either splits the sequence into equal parts at each level of the tree, or splits it into variable-length chunks via an adaptive binding mechanism. All variants of our method are able to substantially outperform prior predictive models that are not goal-conditioned, and the hierarchical variants substantially outperform the sequential one on a long-horizon image-based navigation task. Additionally, the adaptive binding model can discover bottleneck subgoals.
%Our hierarchical GCP planning agent is able to plan faster and more accurately than prior planning methods that employ sequential prediction, and successfully solves complex long-horizon planning tasks.

We present two models for goal-conditioned prediction: a standard sequential architecture and a hierarchical tree-structured variant, where the latter either splits the sequence into equal parts at each level of the tree, or into variable-length chunks via an adaptive binding mechanism. We further propose an efficient hierarchical planning approach based on the tree-structure model. All variants of our method outperform prior video interpolation methods, and the hierarchical variants substantially outperform the sequential model and prior visual MPC approaches on a long-horizon image-based navigation task. Additionally, the adaptive binding model can discover bottleneck subgoals.

\section*{Acknowledgements}
We thank Suraj Nair, Thanard Kurutach, and Aviral Kumar for fruitful discussions. We would like to thank Ben Eysenbach, Ayush Jain and two anonymous internal reviewers for feedback on an earlier version of the paper, Shenghao Zhou for discussion and help with preliminary evaluation of the method, and Kristian Hartikainen for discussion and software development tips.  %This research was supported through the following grants: NSF-IIP-1439681 (I/UCRC), NSF-IIS-1703319, NSF MRI 1626008, ARL RCTA W911NF-10-2-0016, ONR N00014-17-1-2093, ARL DCIST CRA W911NF-17-2-0181, the DARPA-SRC C-BRIC, and by Honda Research Institute.
Support was provided by the ARL DCIST CRA W911NF-17-2-0181 grant, and by Honda Research Institute. KP and OR were visitors at UC Berkeley while conducting this research.

\bibliography{bibref_definitions_long,bibtex}
\clearpage
\appendix

% states when states are avaiable, but observations for the first node in the tree returns the start and the goal observations $\text{pa}(T/2) = (o_1, o_T)$, while for other 

% \label{eq:elbo}
    % \ln p(o_{2:T-1} \vert o_{1,T})   \geq \\
    % \E_{q(z_{2:T-1} \vert x)} \left[\ln p(o_{2:T-1} \vert o_{1,T}, z_{2:T-1})\right] -
    %  \text{KL}\left(q(z \vert o_{1:T}) \;\vert\vert\; p(z_{2:T-1} | o_{1,T})\right). 
% \end{equation}

\section{Additional results}

We include additional qualitative and quantitative results here as well as at the supplementary website: \url{sites.google.com/view/video-gcp}.
\todo{make sure to include standard deviation for all video prediction metrics} %% I don't think that's going to fit in the tables

\begin{figure}[t]
  \centering
  \includegraphics[width=\linewidth]{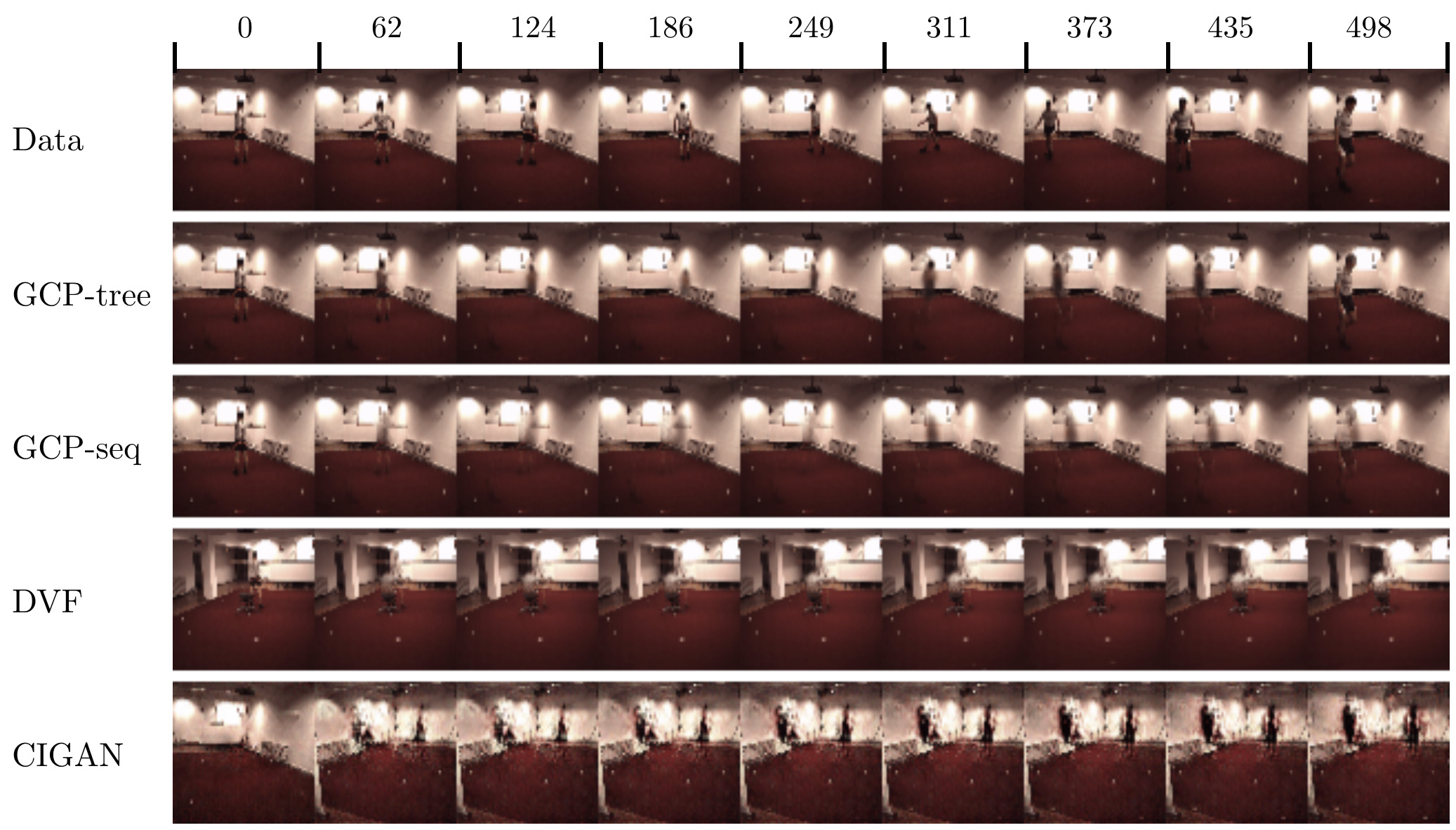}
  \caption{Predictions on Human 3.6M. We see that the GCP models are able to faithfully capture the human trajectory. The optical flow-based method (DVF) captures the background but fails to generate complex motion needed for long-term goal-conditioned prediction. Causal InfoGan also struggles to capture the structure of these long sequences and produce implausible interpolations. Full qualitative results are on the supplementary website: \url{sites.google.com/view/gcp-hier/home}.
  }
  \label{fig:prediction_h36}
\end{figure}

% \begin{figure}
%   \centering
%   \includegraphics[width=\linewidth]{figures/prior_figure.png}
%   \caption{Prior samples from GCP-tree on the four datasets: Human 3.6, \pp, 3x3 Maze and 10x10 Maze. Each sequence is subsampled to 9 frames. Full qualitative results are on the supplementary website: \url{sites.google.com/view/gcp-hier/home}.
%   }
%   \label{fig:prior_samples}
% %   \vskip -0.2in
% \end{figure}

\begin{figure}
    \centering
    \includegraphics[width=1\linewidth]{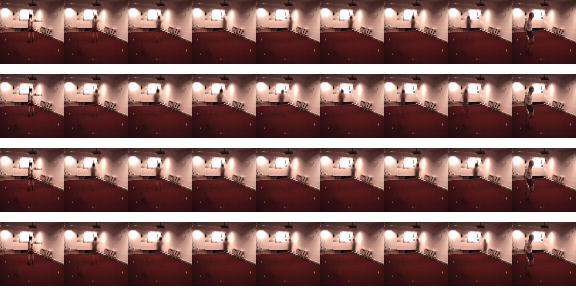}
    \caption{Prior samples from GCP-tree on the Human 3.6M dataset. Each row is a different prior sample conditioned on the same information.}
    \label{fig:prior_h36}
\end{figure}

% \begin{figure}
%     \centering
%     \includegraphics[width=1\linewidth]{figures/maze_multiple_fig.png}
%     \caption{Prior samples from GCP-tree on the 2D maze dataset. Each pair of rows shows two samples given the same context.}
%     \label{fig:prior_h36}
% \end{figure}

\begin{table*}[t]
% \vskip -0.5in
\caption{Prediction performance on perceptual metrics.}
\begin{center}
\begin{footnotesize}
\begin{sc}
    \resizebox{1\textwidth}{!}{
\begin{tabular}{lllllllll}
\toprule
\footnotesize
Dataset  &  \multicolumn{2}{c}{Pick\&Place} &  \multicolumn{2}{c}{Human 3.6M} & \multicolumn{2}{c}{9-room Maze} & \multicolumn{2}{c}{25-room Maze}\\
\cmidrule(r){2-3} \cmidrule(r){4-5} \cmidrule(r){6-7} \cmidrule(r){8-9} 
Method & FVD & LPIPS & FVD & LPIPS & FVD & LPIPS & FVD & LPIPS\\% & PSNR-DTW & SSIM-DTW\\
\midrule

GCP-tree 
& 430.3 & \textbf{0.02}
& \textbf{1314.3} & \textbf{0.05}
& \textbf{655.50} & \textbf{0.174 }
& \textbf{413.31} & \textbf{0.168}
\\
%& \textbf{21.17} & \textbf{0.654} \\

GCP-sequential 
& \textbf{328.9} & \textbf{0.02}
& 1541.8 & \textbf{0.06}
& 860.04 & 0.214 
& 638.95 & 0.238
\\ %$\textbf{472.7} & \textbf{0.26}  \\
%& 20.16 & 0.604 \\

\midrule
DVF  \cite{liu2017video}
& 2879.9 & 0.06  
& 1704.6 & \textbf{0.05} 
& 1320.34 & 0.231
& 1476.44 & 0.215
\\  
%& ? & ? \\

CIGAN \cite{kurutach2018learning}
& 3252.6 & 0.12
& 2528.5 & 0.17 
&  1440.6  & 0.190
& 677.40 & 0.219
\\ 
%& ? & ? \\

% \midrule

% SVG
% & 713.6 & \textbf{0.01}  
% & 1684.7 & 0.08
% & 498.1 & 0.19  
% & 488.5 & 0.28  \\ 
% %& ? & ? \\

\bottomrule
\end{tabular}}
\end{sc}
\end{footnotesize}
\end{center}
\label{tab:app_perceptual_metrics}
% \vskip -0.3in
\end{table*}

\begin{figure}
    \centering
    \includegraphics[width=1\linewidth]{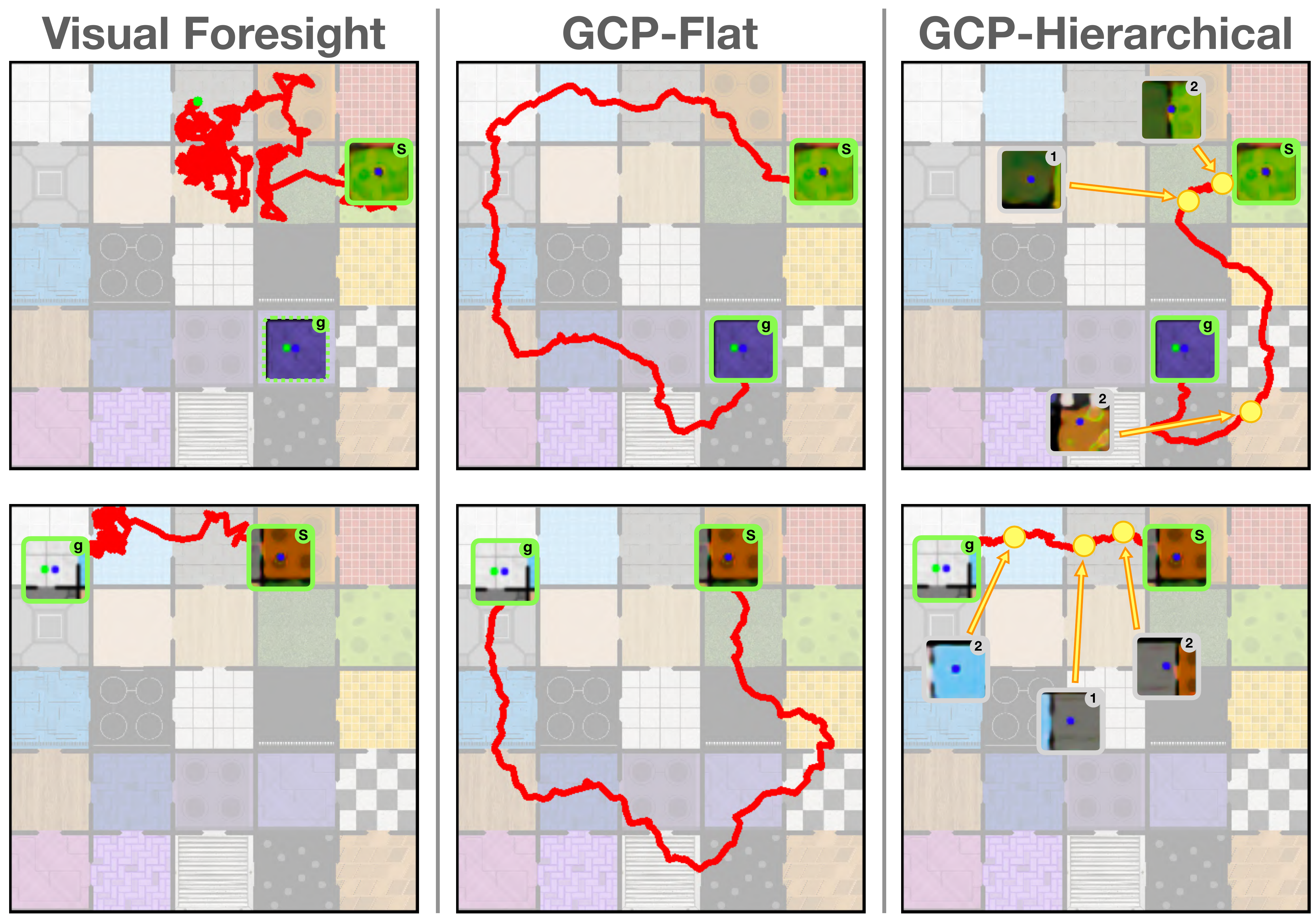}
    \caption{Comparison of visual planning \& control approaches. Execution traces of Visual Foresight (\textbf{left}), GCP-tree with non-hierarchical planning (\textbf{middle}) and GCP-tree with hierarchical planning (\textbf{right}) on two 25-room navigation tasks. Visualized are start and goal observation for all approaches as well as predicted subgoals for hierarchical planning. Both GCP-based approaches can reach faraway goals reliably, but GCP with hierarchical planning finds shorter trajectories to the goal.}
    \label{fig:plan_exec}
\end{figure}

% \begin{figure}
%   \centering
%   \includegraphics[width=\linewidth]{figures/qual_hplan_img.pdf}
%   \caption{
%   Qualitative hierarchical planning results on the navigation task. Our approach is able to iteratively refine the plan and efficiently optimize for a trajectory between start and goal. Note that the examples above also show few planning failure cases, e.g. planned trajectories through a wall due to model inaccuracies (e.g. second-to-last row, right).
%   }
%   \label{fig:qual_hplan}
% \end{figure}

% \begin{figure}
%   \centering
%   \includegraphics[width=0.77\linewidth]{figures/plan_exec_img.png}
%   \caption{
%   Qualitative control rollouts on the navigation task. Goal is depicted as larger green target, start and end point of executed trajectory as blue and green (small) dots respectively. \textbf{Left to right}: ours, goal-conditioned BC, \citet{ebert2018visual}.
%   }
%   \label{fig:qual_control_exec}
% \end{figure}

\section{Evidence lower bound (ELBO) derivation}

We wish to optimize the likelihood of the sequence conditioned on the start and the goal frame $p(o_{2:T-1} \vert o_{1,T})$. However, due to the use of latent variable models, this likelihood is intractable, and we resort to variational inference to optimize it. Specifically, we introduce an approximate posterior network $q(z_{2:T-1} \vert o_{1:T})$, where  that approximates the true posterior \cite{kingma2014auto,rezende2014stochastic}. The ELBO can be derived from the objective that consists of likelihood and a term that enforces that the approximate posterior matches the true posterior: 
\begin{multline}
    \ln p(o_{2:T-1} \vert o_{1,T}) \geq \ln p(o_{2:T-1} \vert o_{1,T}) - \text{KL}(q(z_{2:T-1} \vert o_{1:T}) || p(z_{2:T-1} \vert o_{1:T})) \\ 
     = \E_{q(z_{2:T-1} \vert o_{1:T})} \left[\ln p(o_{2:T-1} \vert o_{1,T}, z_{2:T-1})\right] -
     \text{KL}\left(q(z_{2:T-1} \vert o_{1:T})) \;\vert\vert\; p(z_{2:T-1} | o_{1,T})\right),
\end{multline}
% {align}
% \begin{split}
% \end{split}
% \end{align}
where the last equality is simply a rearrangement of terms. 

Further, in order to efficiently parametrize these distributions, we factorize the distributions as follows according to the graphical model in Fig \ref{fig:gcp_pgm} (right) and Eq. \ref{eq:gcp_tree}:
\begin{align}
    p(o_{2:T-1} \vert o_{1,T}, z_{2:T-1}) & = \prod_{t=2}^{T-1} p(o_{t} \vert o_{1,T}, z_{t}), \\
    p(z_{2:T-1} | o_{1,T}) & = \prod_{t=2}^{T-1},  \\
    q(z_{2:T-1} | o_{1:T}) & = \prod_{t=2}^{T-1} q(z_t | o_t, \text{pa}(t)).
\end{align}
%%OR: note, this derivation is somewhat problematic because it redefines the pa operator. In the main text, it refers to the parent of s, which is only the two immediate parents. Here, it refers to the parents of z, which are all nodes above in the tree (or, alternatively, the immediate parent both h and z). I am not sure how to clarify this

We therefore require the following distributions to define our model: $p(o_{t} \vert o_{1,T}, z_{t})$, $p(z_t | \text{pa}(t))$, $q(z_t | o_t, \text{pa}(t))$. The parameterization of these distributions is defined in Section \ref{sec:arch}. The parent operator $\text{pa}(t)$ returns the parent nodes of $s_t$ according to the graphical model in Fig \ref{fig:gcp_pgm} (right). Using these factorized distributions, we can write out the ELBO in more detail as:
\begin{multline}
    \ln p(o_{2:T-1} \vert o_{1,T}) \geq \E_{q(z_{2:T-1} \vert o_{1:T})} \sum_{t=2}^{T-1} \left[  \ln  p(o_{t} \vert o_{1,T}, z_{t}) -
    \text{KL}\left(q(z_t | o_t, \text{pa}(t))  \;\vert\vert\; p(z_t | \text{pa}(t))\right)  \right].
\end{multline}

\section{Architecture}
\label{sec:app_architecture}

We use a convolutional encoder and decoder similar to the standard DCGAN discriminator and generator architecture respectively. The latent variables $z_n$ as well as $e_n$ are 256-dimensional. All hidden layers in the Multi-Layer Perceptron have 256 neurons. We add skip-connections from the encoder activations from the first image to the decoder for all images. For the inference network we found it beneficial to use a 2-layer 1D temporal convolutional network that adds temporal context into the latent vectors $e_t$. For the recursive predictor that predicts $e_n$, %we found it crucial for the stability of the training to 
we use group normalization \cite{wu2018group}. We found that batch normalization \cite{ioffe2015batch} does not work as well as group normalization for the recursive predictor and conjecture that this is due to the activation distributions being non-i.i.d. for different levels of the tree. We use batch normalization in the convolutional encoder and decoder, and use local per-image batch statistics at test time.  Further, for the simple RNN (without the LSTM architecture) ablation of our tree model, we activate $e_n$ with hyperbolic tangent (tanh). We observed that without this, the magnitude of activations can explode in the lower levels of the tree and conjecture that this is due to recursive application of the same network. %To avoid optimization issues with variational inference, we use the free nats technique with 1 free nat per time step \cite{kingma2016improved}. %Fig.~\ref{fig:architecture} gives a schematic overview of the recursive prediction architecture.
We found that using TreeLSTM \citep{tai2015improved} as the backbone of the hierarchical predictor significantly improved performance over vanilla recurrent architectures. 

To increase the visual fidelity of the generated results when predicting images, we use a foreground-background generation procedure similar to \citep{wang2018video}. The decoding distribution $p(o_t | s_t)$ is a mixture of discretized logistics \cite{salimans2017pixelcnn++}, which we found to work better than alternative distributions. We use the mean of the decoding distribution as the prediction. 

For the adaptive binding model, the frame $o_t$ corresponding to the node $s_n$ is not known before the $s_n$ is produced. We therefore conditioned the inference distribution on the entire evidence sequence $o_{1:T}$ via the attention mechanism over the embeddings \citep{bahdanau2014neural, luong2015effective}: $q(z_t) = \text{Att}(\text{enc}(o_{1:T}), \text{pa}(t))$. We reuse the same observation embeddings $e_t$ for the attention mechanism values. 

The different paths between the same start and goal may have very different lengths (see e.g. Fig. \ref{fig:plan_exec}), so it is necessary for GCP models to predict sequences of different lengths. We do so by training a termination classifier that predicts how long the sequence is. For GCP-Sequential, the termination classifier simply outputs the number of frames in the sequence, and the sequence is produced by recurrently unrolling that many frames. For the GCP-Tree model, to account for varied shapes of the tree, we instead predict a binary termination value at each node. To sample a trajectory, we recursively expand the tree, but stop the expansion where a particular node was classified as terminal (determined by a threshold on the classifier output). This procedure enables us to model even datasets with sequences of variable lengths.

% \todo{Make sure we talk about learned attention}

\paragraph{Hyperparameters.} %For each method and dataset, we performed a manual sweep of the hyperparameter $\beta$ in the range from $1\mathrm{e}{-0}$ to $1\mathrm{e}{-4}$. 
The convolutional encoder and decoder both have five layers. %We  the produced tree after 7 levels (producing a maximum of 127 frames). 
We use the Rectified Adam optimizer \citep{liu2019variance, kingma2014adam} with $\beta_1 = 0.9$ and $\beta_2 = 0.999$, batch size of 16 for GCP-sequential and 4 for GCP-tree, and a learning rate of $\num{2e-4}$. On each dataset, we trained each network for the same number of epochs on a single high-end NVIDIA GPU. Training took a day for all datasets except the 25-room dataset, where we train the models for 3 days. % According to recent research on scaling laws for deep learning we prefer training larger models rather than training models until convergence
% Scaling Laws for Neural Language Models

% \todo{talk about training setup, how long training took / energy consumption}

\section{Data processing and generation}
\label{sec:app_data_maze}
\begin{wrapfigure}{R}{0.4\textwidth}
\vskip -0.2in
  \centering
  \includegraphics[width=\linewidth]{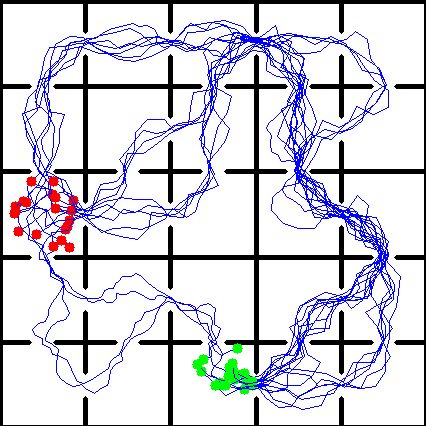}
  \vskip -0.1in
    \caption{Example trajectory distributions between fixed start (red) and goal (green) rooms on the 25-room navigation task. The example goal-reaching behavior is highly suboptimal, with both strong multimodality in the space of possible solutions as well as low-level noise in each individual trajectory.}
    \label{fig:maze_data_dist}
    \vskip -0.4in
\end{wrapfigure}
For training GCPs we use a dataset of example agent goal-reaching behavior. Below we describe how we collect those examples on the \pp and navigation tasks and the details of the Human3.6M dataset. The data can be found on the following links:
\begin{itemize}
    \item 9-room:  \url{https://www.seas.upenn.edu/~oleh/datasets/gcp/nav_9rooms.zip}
    \item 25-room:  \url{https://www.seas.upenn.edu/~oleh/datasets/gcp/nav_25rooms.zip}
    \item \pp:   \url{https://www.seas.upenn.edu/~oleh/datasets/gcp/sawyer.zip}
    \item Pre-processed H3.6:  \url{https://www.seas.upenn.edu/~oleh/datasets/gcp/h36m.zip}
\end{itemize}

\paragraph{\pp.} We generate the \pp dataset using the RoboSuite framework~\cite{corl2018surreal} that is based on the Mujoco physics simulator~\cite{todorov2012mujoco}. We generate example goal-reaching trajectories by placing two objects at random locations on the table and using a rule-based policy to move them into the box that is located at a fixed position on the right of the workspace. We sample the object type randomly from a set of two possible object types, bread and can, with replacement.

\paragraph{Human 3.6M.} For the Human 3.6 dataset, we downsample the original videos to 64 by 64 resolution. %Following prior work, we also temporally subsample the videos with a frequency of 1/8 so that the motion is more pronounced. 
We obtain videos of length of roughly 800 to 1600 frames, which we randomly crop in time to 500-frame sequences. We split the Human 3.6 into training, validation and test set by correspondingly 95\%, 5\% and 5\% of the data. %\KP{Does that follow prior work we can cite for this split? or at least mention that it is standard?} %%OR: no, as far as I understand this is completely made up. %On the TAP dataset, we use 48949 videos for training, 200 for validation and 200 for testing.

\paragraph{Navigation.}  For the navigation task the agent is asked to plan and execute a path between a given 2D start and goal position. The environment is simulated using the Gym-Miniworld framework~\cite{gym_miniworld}. We collect goal-reaching examples by randomly sampling start and goal positions in the 2D maze and plan trajectories using the Probabilistic Roadmap (PRM, \citet{kavraki1996probabilistic}) planner. The navigation problem is designed such that multiple possible room sequences can be traversed to reach from start to goal for any start and goal combination. During planning we sample one possible room sequence at random, but constrain the selection to only such sequences that do not visit any room more than once, i.e. that do not have loops. This together with the random sampling of waypoints of the PRM algorithm leads to collected examples of goal reaching behavior with substantial suboptimality. We show an example trajectory distribution from the data in~Fig.~\ref{fig:maze_data_dist}. While GCPs support training on sequences of variable length we need to set an upper bound on the length of trajectories to bound the required depth of the hierarchical predictive model and allow for efficient batch computation (e.g. at most 200 frames for the 25-room environment). If plans from the PRM planner exceed this threshold we subsample them to the maximum lenght using spline interpolation before executing them in the environment. The training data consists of 10,000 and 23,700 sequences for the 9-room and the 25-room task respectively, which we split at a ration of 99\%, 1\%, 1\% into training, validation and test.

\section{Planning Experimental Setup}
\label{app:sec_planning_details}

For planning with GCPs we use the model architectures described in Section~\ref{sec:app_architecture} trained on the navigation data described in Section~\ref{sec:app_data_maze}. The hyperparameters for the hierarchical planning experiments are listed in Table~\ref{tab:vs_param}. We keep the hyperparameters constant across both 9-room and 25-room tasks except for the maximum episode length which we increase to 400 steps for the 25-room task. Note that the cost function is only used at training time to train the cost estimator described in Section~\ref{sec:planning_control}, which we use to estimate all costs during planning.

To infer the actions necessary to execute a given plan, we train a separate inverse model $a_t = f_{\text{inv}}(o_t, o_{t+1})$ that infers the action $a_t$ which leads from observation $o_t$ to $o_{t+1}$. We train the inverse model with action labels from the training dataset and, in practice, input predicted feature vectors $\hat{e}_t$ instead of the decoded observations to not be affected by potential inaccuracies in the decoding process. We use a simple 3-layer MLP with 128 hidden units in each layer to instantiate $f_{\text{inv}}$. At every time step the current observation along with the next observation from the plan is passed to the inverse model and the predicted action is executed. We found it crucial to perform such closed-loop control to avoid accumulating errors that posed a central problem when inferring the actions for the whole plan once and then executing them open-loop.

We separately tuned the hyperparameters for the visual foresight baseline and found that substantially more samples are required to achieve good performance, even on the shorter 9-room tasks. Specifically, we perform three iterations of CEM with a batch size of 500 samples each. For sampling and refitting of action distributions we follow the procedure described in~\cite{nagabandi2019deep}. We use a planning horizon of 50 steps and replan after the current plan is executed. We cannot use the cost function from Table~\ref{tab:vs_param} for this baseline as it leads to degenerate solutions: in constrast to GCPs, VF searches over the space of \emph{all} trajectories, not only those that reach the goal. Therefore, the VF planner could minimize the trajectory length cost used for the GCP models by predicting trajectories in which the agent does not move. We instead use a cost function that measures whether the predicted trajectory reached the goal by computing the L2 distance between the final predicted observation of the trajectory and the goal observation.

We run all experiments on a single NVIDIA V100 GPU and find that we need approximately 30mins / 1h to evaluate all 100 task instances on the 9-room and 25-room tasks respectively when using the hierarchical GCP planning. The VF evaluation requires many more model rollouts and therefore increases the runtime by a factor of approximately five, even though we increase the model rollout batch size by a factor of 20 for VF to parallelize trajectory sampling as much as possible.

\begin{table}
\caption{Hyperparameters for hierarchical planning with GCPs on 9-room and 25-room navigation tasks.}
  \centering
  \begin{tabular}{ll}
    \toprule
    \multicolumn{2}{c}{Hierarchical Planning Parameters}\\
    \midrule
    Hierarchical planning layers ($D$) & 2  \\
    Samples per subgoal ($M$) & 10  \\
    \midrule
    \multicolumn{2}{c}{Final Segment Optimization}\\
    \midrule
    Sequence samples per Segment & 5 \\
    \midrule
    \multicolumn{2}{c}{General Parameters}\\
    \midrule
    Max. episode steps & 200 / 400 \\
    Cost function & $\sum_{t=0}^{T-1} (x_{t+1} - x_t)^2$ \\
    \bottomrule
  \end{tabular} 
\label{tab:vs_param}
\end{table}

\section{Adaptive Binding with Dynamic Programming}
\label{seq:hedge}

\subsection{An efficient inference procedure}

To optimize the model with adaptive binding, we perform variational inference on both $w$ and $z$:

\begin{equation}
\log p (x) \geq  \mathbb{E}_{q(z,w)} [p(x | w,z)] - D_{KL} (q(z|x) || p(z)) - D_{KL} (q(w|x,z) || p(w)).
\end{equation}

To infer $q(w|x,z)$, we want to produce a distribution over possible alignments between the tree and the evidence sequence. Moreover, certain alignments, such as the ones that violate the ordering of the sequence are forbidden. We define such distribution over aligment matrices $A$ via Dynamic Time Warping. We define the energy of an alignment matrix as the cost, and the following distribution over alignment matrices: 
$$ p(A|x,z) = \frac{1}{Z} e^{-A * c(x,z)},$$
where the partition function $Z = \mathbb{E}_A [e^{-A * c(x,z)}]$, and $c$ is the MSE error between the ground truth frame $x_t$ and the decoded frame associated with $z_n$. We are interested in computing marginal edge distributions $w = \mathbb{E}_A [A]$. Given these, we can compute the reconstruction error efficiently. We next show how to efficiently compute the marginal edge distributions.

Given two sequences $x_{0:T}, z_{0:N}$, denote the partition function of aligning two subsequences $x_{0:i}, z_{0:j}$ as $f_{i,j} = \sum_{A \in \mathcal{A}_{0:i, 0:j}} e^{-A * c(x_{0:i}, z_{0:j})}$. \cite{cuturi2017soft} shows that these can be computed efficiently as:
$$ f_{i,j} = c(x_i, z_j) * (f_{i-1, j-1} + f_{i-1, j}).$$
We note that we do not include the third term $f_{i, j-1})$, as we do not want a single predicted frame to match multiple ground truth frames. Furthermore, denote the partition function of aligning $x_{i:T}, z_{j:N}$ as $b_{i,j} = \sum_{A \in \mathcal{A}_{i:T, j:N}} e^{-A * c(x_{i:T}, z_{j:N})}$. Analogously, we can compute it as: 
$$ b_{i,j} = c(x_i, z_j) * (b_{i+1, j+1} + b_{i+1, j}).$$

\begin{prop}
The total unnormalized density of all alignment matrices that include the edge $(i,j)$ can be computed as $e_{i,j} = f_{i,j} * b_{i,j} \slash c(x_i, z_j) =  c(x_i, z_j) * (f_{i-1, j-1} + f_{i-1, j}) * (b_{i+1, j+1} + b_{i+1, j}) $. Moreover, the probability of the edge $(i,j)$ can be computed as $w_{i,j} = e_{i,j} / Z$.
\end{prop}

Proposition 1 enables us to compute the expected reconstruction loss in quadratic time: $$p(x|z) = w * c(x,z).$$

\subsection{Bottleneck Discovery Experimental Setup}

In order to use the adaptive binding model to discover bottleneck frames that are easier to predict, we increase the reconstruction loss on those nodes as described in the main text. Specifically, we use Gaussian decoding distribution for this experiment, and set the variance of the decoding distribution for several top layers in the hierarchy to a fraction of the value for lower layers. This encourages the model to bind the frames that are easier to predict higher in the hierarchy as the low variance severely penalizes poor predictions. We found this simple variance re-weighting scheme effective at discovering bottleneck frames on several environments. %\todo{adapt this to the explanation with chaning KL coefficients we use in the paper. add any relevant hyperparameter values that differ from the original PP experiments (maybe regrading attention mechanism etc that was not used with the balanced model)}

To generate the visualization of the discovered tree structure in Fig.~\ref{fig:bottleneck} we evenly subsample the original 80-frame sequences and display those nodes that bound closest to the subsampled frames such that the resulting graph structure still forms a valid 2-connected tree. The variations in tree structure arise because the semantic bottlenecks which the nodes specialize on binding to appear at different time steps in the sequences due to variations in speed and initial position of the robot arm as well as initial placement of the objects.

\section{Training from Random Data}
\label{sec:random_data}

\begin{wraptable}{R}{0.6\linewidth}
% \vskip -0.25in
\caption{Average Trajectory Length. Planning with GCP finds shorter paths than the training distribution.} \label{tab:traj_lenth}
\begin{center}
% \begin{small}
\begin{sc}
    % \resizebox{1\textwidth}{!}{
% \vskip -0.15in
% \resizebox{1\linewidth}{!}{
\begin{tabular}{lcc}
\toprule
 & Original Data & Random Data \\
\midrule
Training Data   & $31.4$  &  $62.6$  \\
GCP-Tree (ours)   & $\bm{20.7}$  &  $\bm{42.6}$  \\
\bottomrule
\end{tabular}
% }
\end{sc}
% \end{small}
\end{center}
% \vskip -0.2in
\end{wraptable}

\begin{wrapfigure}{R}{0.45\textwidth}
% \vspace{-15pt}
    \centering
    \includegraphics[width=\linewidth]{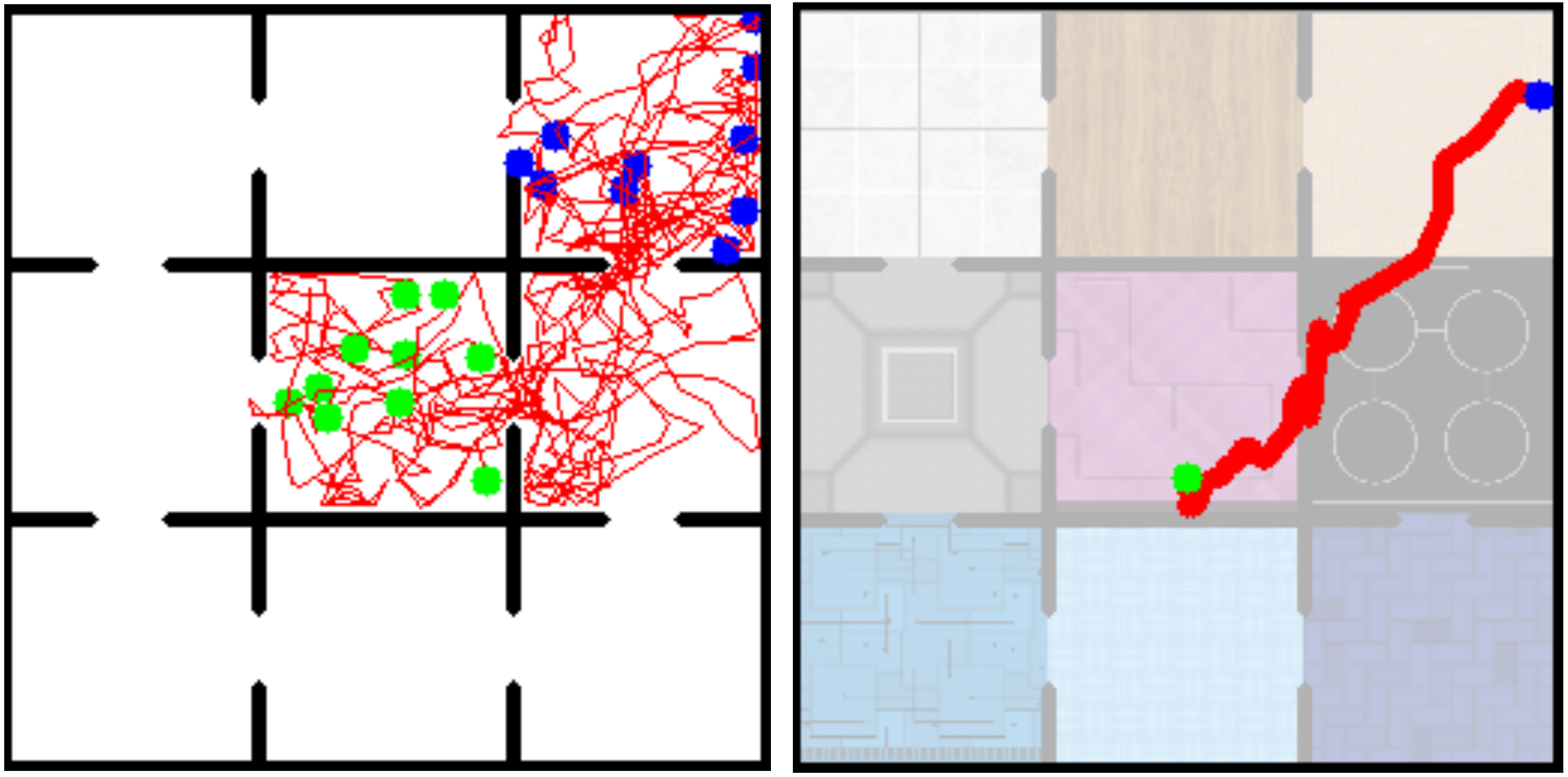}
    \vspace{-15pt}
    \caption{\textbf{Left}: random exploration data. \textbf{Right}: execution of our method trained on random data. %Our method can be trained from data collected by executing random actions, generalize to new start-goal combinations and find shorter paths than seen during training.
    }
    \label{fig:random_quali_plan}
    % \vspace{-10pt}
\end{wrapfigure}

In the room navigation experiments we train our model with noisy trajectories that reach diverse goals with considerable suboptimality (see Fig.~\ref{fig:maze_data_dist}). To test whether our method can work with even more suboptimal training data, we conduct preliminary experiments with completely random exploration data, and observe that our method still successfully solves navigation tasks in the 9-room environment (see Fig.~\ref{fig:random_quali_plan}). This suggests that the proposed method is scalable even to situations where no good planners exist that can be used for data collection.

In Tab.~\ref{tab:traj_lenth}, we compare the average trajectory length of training data and our method on both, the dataset used for the experiments in section~\ref{sec:video_pred} and the random action data. We find that planning with our method leads to substantially shorter trajectories, further showing the ability of our approach to improve upon low-quality training data.

\subsection{Runtime Complexity}

\paragraph{Computational efficiency.} While the sequential forward predictor performs $\gO(T)$ sequential operations to produce a sequence of length T, the hierarchical prediction can be more efficient due to parallelization. As the depth of the tree is $\ceil{\log T}$, it only requires $\gO(\log T)$ sequential operations to produce a sequence, assuming all operations that can be conducted in parallel are parallelized perfectly. We therefore batch the branches of the tree and process them in parallel at every level to utilize the benefit of efficient computation on modern GPUs. We note that the benefits of the GCP-tree runtime lie in parallelization, and thus diminish with large batch sizes, where the parallel processing capacity of the GPU is already fully utilized. We notice that, when predicting video sequences of 500 frames, GCP-sequential can use up to 4 times bigger batches than GCP-Tree without significant increase in runtime cost. This benefit is applicable both during training and inference. \todo{we might just want to cut this part} %\KP{this whole paragraph could move to the experimental section, esp the last sentence} \JD{Agreed with Karl.}
%DJ9.23: Some of this seems like it belongs in results rather than in approach.
%DJ.5.25: Does GCP-tree still have significant computational gains over GCP-hierarchical at test time?
%OR.5.26: Added a sentence about this. We don't have a killer app for this at test time, but it would matter if the planning needed more samples

When training tree-structured networks we exploit the provided parallelism in the structure of the model and batch recursions in the tree that are independent when conditioned on their parents.

\end{document}